\documentclass[11pt]{article}

\usepackage{acl}

\usepackage{times}
\usepackage{latexsym}

\usepackage[T1]{fontenc}

\usepackage[utf8]{inputenc}

\usepackage{microtype}
\usepackage{inconsolata}
\usepackage{graphicx}
\usepackage{enumitem}
\usepackage{multirow}
\usepackage{booktabs}
\usepackage[table]{xcolor}
\usepackage{amsmath,amssymb,amsfonts}
\usepackage{algorithm}
\usepackage{algpseudocode}
\algrenewcommand{\algorithmiccomment}[1]{\hfill{\color{gray!85}\textit{\%\ #1}}}

\title{RVSD: Retrieval Vision Sparse Decoding for Mitigating Visual Hallucinations in Large Vision-Language Models}

\author{
  \setcounter{footnote}{1}%
  Canjie Liu$^{1}$, Jiawen Kang\thanks{\, Corresponding author.}$^{1}$, Jinbo Wen$^{2}$, Zishao Zhong$^{3}$ \\
  \normalfont\normalsize $^{1}$\, School of Automation, Guangdong University of Technology, Guangzhou, China \\
  \normalfont\normalsize $^{2}$\, Department of Computer Science, City University of Hong Kong, Hong Kong, China \\
  \normalfont\normalsize $^{3}$\, The Second Affiliated Hospital of Guangzhou University of Chinese Medicine, Guangzhou, China \\
  \normalfont\normalsize \texttt{2112404359@mail2.gdut.edu.cn}, \texttt{kavinkang@gdut.edu.cn} \\
  \normalfont\normalsize \texttt{jinbowen2-c@my.cityu.edu.hk}, \texttt{zhongzishao@gzucm.edu.cn}
}

\begin{document}
\maketitle

\begin{abstract}
Large vision-language models have achieved remarkable success in vision-language tasks. However, they remain prone to Visual Hallucinations (VHs), undermining their reliability in real-world applications. Existing solutions typically require curated datasets, additional training, or multi-round decoding, resulting in considerable computational overhead. In this paper, we propose \textbf{RVSD} (\underline{R}etrieval \underline{V}ision \underline{S}parse \underline{D}ecoding), a training-free and plug-and-play decoding framework that, for the first time, unifies token sparsification and \textbf{Semantic-Space Visual Retrieval} (SSVR) within a single decoding pass. Within RVSD, we introduce a \textbf{semantics-directed token selection} strategy that selectively sparsifies redundant tokens while preserving critical visual information. We further propose the SSVR mechanism, which reformulates visual compensation as an on-demand cross-modal retrieval process within a shared semantic space. Extensive experiments demonstrate that RVSD achieves state-of-the-art performance in mitigating VHs while maintaining robust suppression capabilities under long-context generation settings. Our code is available here.\footnote{https://github.com/canjie-liu/RVSD}
\end{abstract}

\section{Introduction}
\begin{figure}[!t]
    \centering
    \includegraphics[width=\columnwidth]{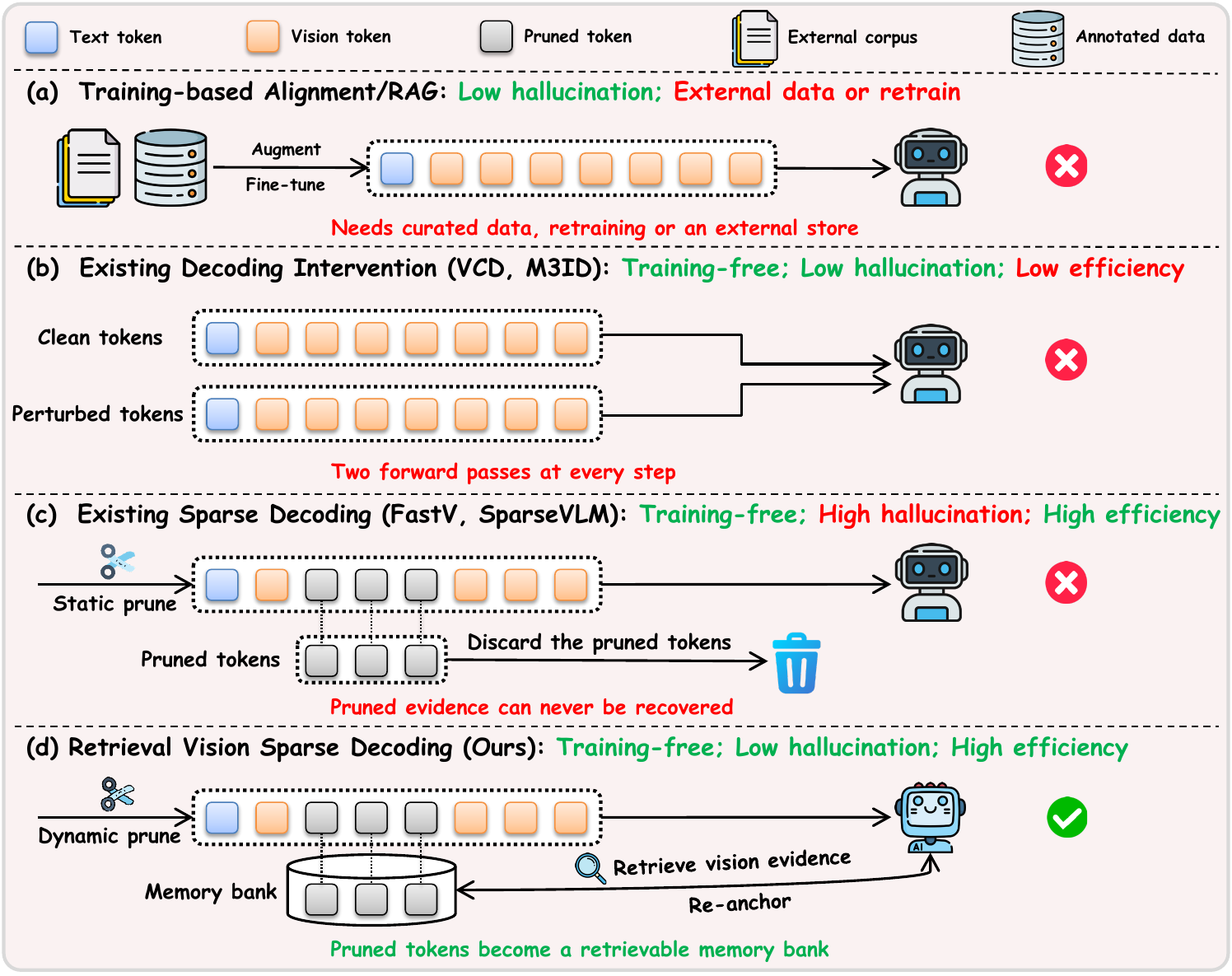}
    \captionsetup{font=small}
    \caption{\label{fig:motivation} Comparison of representative paradigms for mitigating VHs in LVLMs: (a) training-based alignment and RAG~\citep{sun2024aligning}, (b) decoding interventions~\citep{leng2024mitigating,favero2024multi}, (c) sparse decoders~\citep{chen2024image,pmlr-v267-zhang25s}, and (d) our RVSD, which turns pruned tokens into a retrievable memory bank and restores grounding on demand in a single forward pass.}
\end{figure}

Large Vision-Language Models (LVLMs) have recently demonstrated impressive capabilities across a wide range of vision-language tasks~\citep{liu2024improved,bai2023qwen,liu2024llavanext}, such as image captioning, visual question answering, and open-ended visual reasoning. Despite their rapid progress, LVLMs remain highly susceptible to Visual Hallucinations (VHs)~\citep{bai2024hallucination, liu2024survey}, where generated responses contradict or lack grounding in the actual visual content. This fundamental flaw seriously compromises the reliability of LVLMs in real-world applications~\citep{wang2023chatcad}, where misinterpretation of visual data can lead to significant consequences.

Existing solutions can be categorized into four paradigms, which we compare with RVSD in Fig.~\ref{fig:motivation}. Training-based alignment~\citep{sun2024aligning} mitigates VHs via additional supervision and preference alignment, but requires expensive retraining and annotated data. Retrieval-augmented generation~\citep{lewis2020retrieval} grounds generation using external knowledge bases, yet introduces substantial storage and retrieval overhead. Decoding-side interventions~\citep{liu2025reducing} adjust token distributions during inference, but often require iterative decoding procedures that incur considerable latency. More recently,
sparsification-based decoding~\citep{dang2026locate} has emerged as a promising direction that simultaneously accelerates inference and suppresses distracting visual tokens, offering an attractive trade-off between efficiency and VH mitigation for LVLMs.

\begin{figure}[t]
\centering
\includegraphics[width=0.45\textwidth]{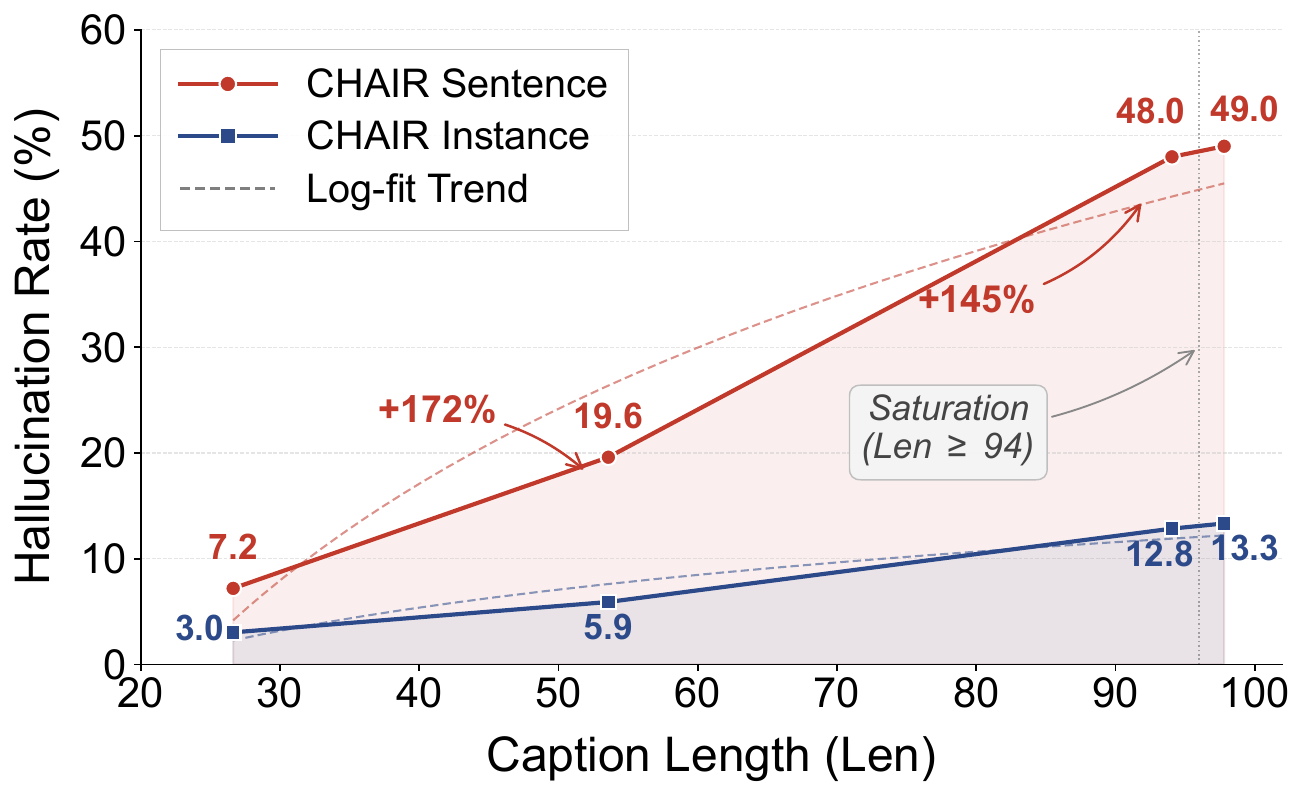}
\captionsetup{font=small}
\caption{\label{fig:len_chair_growth} Sparsification-hallucination paradox, exemplified by the representative sparse decoder VASparse~\citep{zhuang2025vasparse} on LLaVA-1.5 using the Caption Hallucination Assessment with Image Relevance (CHAIR) dataset.}
\end{figure}

However, existing sparse decoding methods remain largely modality-agnostic, often pruning visual tokens that appear statistically redundant yet contain semantically critical information~\citep{zhuang2025vasparse}. Once such fine-grained visual tokens are discarded, the resulting information loss becomes irreversible. Our empirical analysis further reveals that the hallucination rates of State-of-the-Art (SOTA) sparse decoders~\citep{zhuang2025vasparse} increase substantially as generation length grows, as shown in Fig.~\ref{fig:len_chair_growth}. We refer to this phenomenon as the \textit{sparsification-hallucination paradox}, highlighting the fundamental tension between inference efficiency and reliable visual grounding in LVLMs.

In this paper, we propose \textbf{RVSD} (\underline{R}etrieval \underline{V}ision \underline{S}parse \underline{D}ecoding), a training-free and plug-and-play decoding framework for mitigating VHs in LVLMs. RVSD addresses the sparsification-hallucination paradox by combining semantics-directed token sparsification with on-demand visual compensation in a unified single-pass inference process. Our contributions are as follows:
\begin{itemize}
  \item We propose RVSD, a novel training-free and plug-and-play decoding framework that addresses the sparsification-hallucination paradox. RVSD jointly integrates token sparsification and dynamic visual compensation within a single inference pass, achieving an effective balance between inference efficiency and visual faithfulness.
  \item We develop a semantics-directed token selection strategy that selectively distinguishes redundant tokens from visually critical ones. Unlike prior modality-agnostic pruning methods, the proposed strategy performs targeted sparsification while preserving essential visual grounding information.
  \item We propose a Semantic-Space Visual Retrieval (SSVR) mechanism that performs on-demand cross-modal retrieval in a shared semantic space. When visual grounding becomes insufficient during generation, SSVR dynamically retrieves relevant visual evidence from a deferred visual memory bank to recalibrate hallucinated output distributions.
  \item Extensive experiments across multiple prominent benchmarks demonstrate that RVSD achieves SOTA performance, simultaneously improving both generation quality and decoding efficiency while exhibiting strong robustness under long-context generation settings.
\end{itemize}

\begin{figure*}[t]
    \centering
    \captionsetup{font=small}
    \includegraphics[width=0.95\textwidth]{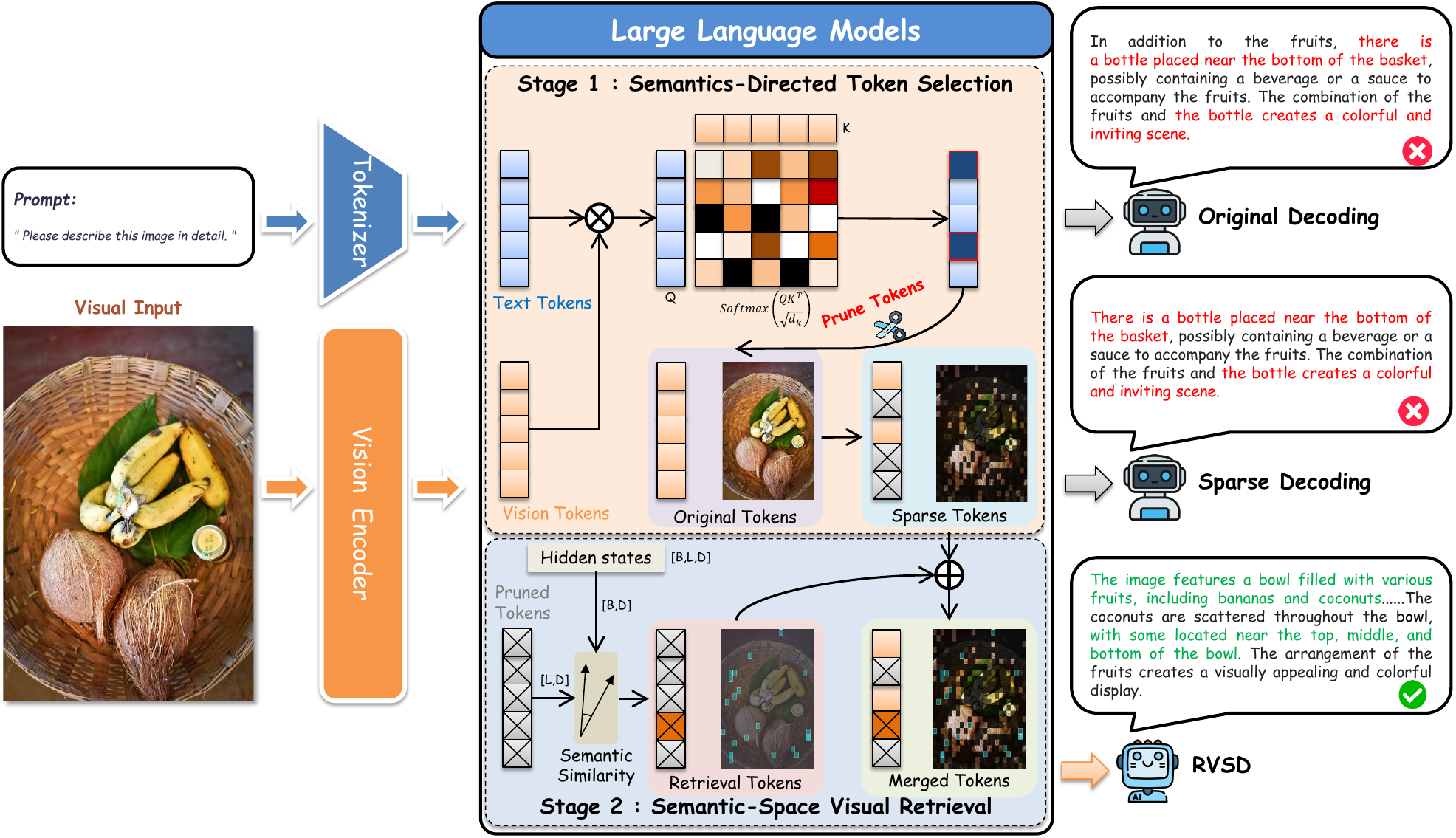}
    \caption{\label{fig:rvsd_overview} Overview of the proposed RVSD framework, which consists of two stages. Stage 1 performs semantics-directed token selection, partitioning visual tokens into an active sparse set and a deferred memory bank through cross-modal attention. Stage 2 applies the SSVR mechanism to retrieve on-demand visual evidence from the deferred bank and re-anchor decoding within a single forward pass, transforming hallucinated content from the original and sparse decoders into faithful generation.}
\end{figure*}

\section{Related Work}

\noindent\textbf{Large Vision-Language Models.}
Building upon the rapid advances in large language models~\citep{zhao2026survey}, recent research has extended them to the multimodal domain, giving rise to LVLMs~\citep{liu2023visual,liu2024improved,liu2024llavanext}. LVLMs are typically trained via large-scale vision-language pretraining followed by instruction-based fine-tuning, while subsequent studies further align generation behavior with human preferences via Reinforcement Learning from Human Feedback (RLHF)~\citep{sun2024aligning} and preference optimization techniques. Despite their strong multimodal capabilities, LVLMs remain prone to VHs~\citep{bai2024hallucination, Yin_2024, wu2023multimodal}, severely undermining the reliability of LVLMs in real-world applications.

\smallskip\noindent\textbf{Visual Hallucination Mitigation.}
Existing efforts mitigate VHs from several perspectives. Training-centric approaches~\citep{sun2024aligning} improve multimodal alignment through additional supervision, RLHF, or preference optimization techniques. Decoding-time methods intervene during inference, where contrastive strategies, such as VCD~\citep{leng2024mitigating} and M3ID~\citep{favero2024multi}, suppress linguistic priors via perturbed visual inputs, while attention-guided methods, including VTI~\citep{liu2025reducing}, AvisC~\citep{woo2025don}, and SHARP~\citep{wu2025sharp}, enhance attention toward visually relevant regions. Adaptive visual reinforcement methods~\citep{zhu2026look} further strengthen visual evidence on demand during generation. Closest to our setting, RVCD~\citep{lee2025retrieval} performs retrieval-based contrastive decoding by referencing an offline database of generated single-concept images at the logit level. In contrast, RVSD requires no external image store, detector, or draft decoding pass, and retrieves the model's own deferred visual tokens directly in the semantic space of a single forward pass. Despite their different formulations, these methods still rely on dense visual processing throughout decoding, leaving substantial room for improving efficiency-aware visual grounding.

\smallskip\noindent\textbf{Sparse Decoding for Efficient Inference.}
Token sparsification has been utilized to improve both inference efficiency and visual faithfulness~\citep{wu2026accelerating}. General-purpose accelerators, such as SparseVLM~\citep{pmlr-v267-zhang25s} and FastV~\citep{chen2024image}, prune visual tokens using modality-agnostic criteria primarily designed for acceleration. Recent methods, such as VASparse~\citep{zhuang2025vasparse}, LTS-FS~\citep{dang2026locate}, PruneHal~\citep{sun2025prunehal}, MINT~\citep{wang2025mint}, and EAZY~\citep{che2025hallucinatory}, further explore hallucination-aware sparsification by enabling LVLMs to focus on task-relevant visual tokens. Nevertheless, existing methods still perform irreversible pruning, which may remove fine-grained visual evidence. Unlike prior methods, RVSD reorganizes discarded tokens into a deferred visual memory bank, enabling on-demand restoration of visual grounding within the same decoding pass without additional training or iterative decoding.

\section{Problem Formulation}

Given an LVLM $\mathcal{M}_{\theta}^{\mathrm{MLLM}}$ parameterized by $\theta$, the model consists of a text embedding layer, a vision encoder, a vision-language interface module, a decoder with $L$ transformer layers, and an affine output layer $\varsigma(\cdot)$ for next-token prediction.
For an image-grounded generation task with textual query $x$ and input image $v$, the vision encoder first extracts visual features from $v$, which are then transformed into aligned visual tokens $z_v$ via a Multi-Layer Perceptron (MLP) or Q-Former interface. The visual tokens $z_v$ are concatenated with the query $x$ and fed into the decoder to autoregressively generate the response $y$, given by
\begin{equation}\label{Auto_generated}
\small
y_t \sim p_{\theta}(\cdot | v, x, y_{<t}) \propto \mathrm{softmax}( f_{\theta}(\cdot | v, x, y_{<t})),
\end{equation}
where $y_t$ denotes the token generated at step $t$, $y_{<t}$ denotes the previously generated prefix, and $f_{\theta}$ represents the unnormalized logits from $\mathcal{M}_{\theta}^{\mathrm{MLLM}}$.

When the generated response $y$ contradicts or is unsupported by the input image $v$, the LVLM is considered to exhibit VHs. Accordingly, the objective of VH mitigation is to suppress unfaithful generations while preserving overall response quality. In the following, we present RVSD, which improves visual faithfulness during inference with minimal computational overhead.

\section{Methodology}

As illustrated in Fig.~\ref{fig:rvsd_overview}, we present the architecture of the proposed RVSD. In the following, we describe its single-pass inference process and two key components: semantics-directed token selection and the SSVR mechanism.

\subsection{Single Forward Pass}
Let the visual token bank be denoted by $Z_v=[z_1,\ldots,z_{N_v}] \in \mathbb{R}^{N_v \times D}$, where $N_v$ is the total number of visual tokens and $D$ is the hidden dimension. The decoder hidden states at decoding step $t$ in layer $l$ are denoted by $H_t^{(l)}=[h_1^{(l)},\ldots,h_t^{(l)}]$, in which each $h_i^{(l)}\in\mathbb{R}^{D}$ denotes the hidden state at textual position $i$. Thus, the text-to-vision attention weight $A_{i,j}^{(l)}$ from the $i$-th text token to the $j$-th visual token at layer $l$ is defined as
\begin{equation}
\small
A_{i,j}^{(l)}=
\frac{\exp\left((W_Q^{(l)}h_i^{(l)})^{\top}(W_K^{(l)}z_j)/\sqrt{D}\right)}
{\sum_{r=1}^{N_v}\exp\left((W_Q^{(l)}h_i^{(l)})^{\top}(W_K^{(l)}z_r)/\sqrt{D}\right)},
\end{equation}
where $W_Q^{(l)},W_K^{(l)}$ denote the query and key projection matrices at layer $l$, respectively.

At each decoding step, RVSD maintains two complementary visual token sets: an active sparse set $\mathcal{S}_t$ for standard decoding and a deferred set $\mathcal{P}_t$ that forms a deferred visual memory bank, where $\mathcal{S}_t \cup \mathcal{P}_t={1,\ldots, N_v}$ and $\mathcal{S}_t \cap \mathcal{P}_t=\varnothing$. Decoding is performed over $\mathcal{S}_t$, while $\mathcal{P}_t$ remains semantically retrievable and provides relevant visual evidence on demand. Based on Eq.~(\ref{Auto_generated}), RVSD autoregressively predicts the token at generation step $t$ as
\begin{equation}\label{eq:rvsd_predict}
\small
p_{\theta}^{\mathrm{RVSD}}=\mathrm{softmax}\left(f_{\theta}(\cdot| Z_v[\mathcal{S}_t],x,y_{<t};\Delta_t)\right),
\end{equation}
where $\Delta_t$ denotes an optional SSVR branch activated only when visual grounding becomes insufficient. This design preserves decoding efficiency under normal generation while enabling query-driven visual retrieval during uncertain steps, without requiring multi-round decoding. Algorithm~\ref{alg:rvsd} presents the overall single forward pass of RVSD. The dynamic sparsity module for creating $\mathcal{S}_t$ and $\mathcal{P}_t$ and the SSVR module realizing $\Delta_t$ are detailed in Sections~\ref{sec:dyn_sparse} and~\ref{sec:ssvr}, respectively.

\begin{algorithm}[t]
\captionsetup{font=small}
\caption{Single Forward Pass of RVSD}
\label{alg:rvsd}
\small
\begin{algorithmic}[1]
\Require An LVLM $\mathcal{M}_\theta$, an image $v$, and a query $x$
\State Encode visual tokens: $Z_v \leftarrow \mathcal{M}_\theta^{\mathrm{enc}}(v)$; initialize $\tilde{s}^{0} \leftarrow \mathbf{0}$

\For{$t = 1, 2, \ldots$}
    \State Select token subset $\mathcal{T}_t$ via $\beta^t$ using Eq.~(\Ref{eq:saliency})
    \State Update smoothed saliency $\tilde{s}^t$ from $s^t$ using Eq.~(\Ref{eq:smooth})
    \State Partition visual tokens into $\mathcal{S}_t$ and $\mathcal{P}_t$ using Eq.~(\Ref{eq:sparse_set})
    \State $\textit{trigger} \leftarrow \textsc{True}$

    \For{$l = l_{\mathrm{s}}$ \textbf{to} $l_{\mathrm{e}}$}
        \State Compute uncertainty score $u_t^{(l)}$ using Eq.~(\Ref{eq:uncertainty})

        \If{\textit{trigger} \textbf{and} $u_t^{(l)} > \gamma$}
            \State Retrieve deferred tokens from $Z_v[\mathcal{P}_t]$ via query $q_t^{(l)}$ using Eq.~(\Ref{eq:retrieval})
            \State Aggregate retrieved features $\tilde{Z}_v$ from top-$k_c$ candidates using Eq.~(\Ref{eq:aggregate})
            \State Construct adaptive adapter weights $W_{1,2}^{\mathrm{adpt}}$ using Eq.~(\Ref{eq:adapter_w})
            \State Fuse retrieved visual evidence using Eq.~(\Ref{eq:fusion})
            \State $\textit{trigger} \leftarrow \textsc{False}$
        \EndIf
    \EndFor

    \State Predict token $y_t$ over $Z_v[\mathcal{S}_t]$ using Eq.~(\Ref{eq:rvsd_predict})
\EndFor

\State \Return Response sequence $y$
\end{algorithmic}
\end{algorithm}

\subsection{Dynamic Sparsity}
\label{sec:dyn_sparse}

The objective of dynamic sparsity is to retain visual tokens most relevant to the salient textual semantics in the current generation trajectory. Rather than performing modality-agnostic pruning, RVSD explicitly scores visual tokens by aggregating cross-modal attention from salient textual positions. At decoding step $t$, we first compute textual saliency over generated tokens $\{1,\ldots,t-1\}$ as
\begin{equation}\label{eq:saliency}
\small
\beta_i^t=
\frac{\exp(\|h_i^{(L)}\|_2/\tau)}
{\sum_{r=1}^{t-1}\exp(\|h_r^{(L)}\|_2/\tau)},
\end{equation}
where $\|\cdot\|_2$ denotes the $\ell_2$ norm, $h_i^{(L)}$ is the last-layer hidden state of the $i$-th token, and $\tau$ is a temperature parameter.
We then select the major textual token set $\mathcal{T}_t=\mathrm{TopK}(\beta^t,m)$, where $m$ is the number of major textual tokens retained.
Given a layer subset $\mathcal{L}_s$ for stable attention aggregation, we define the visual relevance score $s_j^t$ as
\begin{equation}\label{eq:relevance}
\small
s_j^t=\sum_{i\in\mathcal{T}_t}\beta_i^t\Big(\frac{1}{|\mathcal{L}_s|}\sum_{l\in\mathcal{L}_s}A_{i,j}^{(l)}\Big).
\end{equation}

To suppress step-wise noise, we apply temporal smoothing, which is expressed as
\begin{equation}\label{eq:smooth}
\small
\tilde{s}^t=\eta\tilde{s}^{t-1}+(1-\eta)s^t,
\end{equation}
where $\eta\in[0,1)$ is the smoothing coefficient, and $\tilde{s}^t$ represents the smoothed relevance vector at step $t$. We then construct the active sparse set as
\begin{equation}\label{eq:sparse_set}
\small
\mathcal{S}_t=\mathrm{TopK}(\tilde{s}^t,k),
\:\:
\mathcal{P}_t=\{1,\ldots,N_v\}\setminus\mathcal{S}_t,
\end{equation}
where $k$ is the budget that controls the size of the retained sparse set. Therefore, RVSD retains the visual tokens most relevant to the current decoding context while deferring the remaining tokens for potential later retrieval. With $k \ll N_v$, the per-layer attention complexity is reduced from $\mathcal{O}((t+N_v)^2)$ to $\mathcal{O}((t+k)^2)$.

\subsection{Semantic-Space Visual Retrieval}
\label{sec:ssvr}

Dynamic sparsity alone may still discard fine-grained visual evidence required at later decoding stages. To address this issue, RVSD reformulates visual compensation as on-demand cross-modal retrieval in a shared semantic space. Instead of directly restoring pruned features, RVSD treats the deferred set $\mathcal{P}_t$ as a deferred visual memory bank and uses the current decoder state as a semantic query to retrieve relevant visual evidence whenever grounding becomes insufficient.

\subsubsection{Uncertainty-triggered retrieval gate} At decoding step $t$, we scan intermediate layers and compute the normalized predictive uncertainty as
\begin{equation}\label{eq:uncertainty}
\small
u_t^{(l)}=-\frac{1}{\log N}\sum_{n=1}^{N}p_{\theta,n}^{(l)}\log p_{\theta,n}^{(l)},
\end{equation}
where $N$ is the vocabulary size, and $p_{\theta,n}^{(l)}$ represents the predicted probability of the $n$-th vocabulary token derived from the layer-$l$ hidden state. When $u_t^{(l)}>\gamma$ for $l \in [l_{\mathrm{s}}, l_{\mathrm{e}}]$, where $\gamma$ is a predefined entropy threshold and $[l_{\mathrm{s}}, l_{\mathrm{e}}]$ denotes the intermediate-layer scanning window, the model triggers a single retrieval call at this step, indicating that the current generation trajectory has insufficient visual grounding and requires re-anchoring.

\subsubsection{Cross-modal retrieval}
Once triggered, we project the current decoder hidden state into the shared visual-text semantic space and use it as a query $q_t^{(l)}=h_t^{(l)}[-1,:]$, i.e., the hidden state at the last token position of layer $l$, to query the deferred visual memory bank:
\begin{equation}\label{eq:retrieval}
\small
r_j=\frac{q_t^{(l)}z_j^{\top}}{\sqrt{D}},\; j\in\mathcal{P}_t,
\:\:
w=\mathrm{softmax}(r),
\end{equation}
where $r_j$ measures the semantic relevance between the textual query and each deferred visual token, and $w$ defines a normalized retrieval distribution over $\mathcal{P}_t$. With $k_c=\min(K,|\mathcal{P}_t|)$, where $K$ is a predefined upper bound, we select the top-$k_c$ tokens $(\tilde{w},\mathcal{I}_t)=\mathrm{TopK}(w,k_c)$, where $\tilde{w}$ denotes the corresponding weights, and $\mathcal{I}_t \subseteq \mathcal{P}_t$ denotes their indices in the deferred memory bank. The weights are renormalized and aggregated into a query-conditioned visual representation, which is expressed as
\begin{equation}\label{eq:aggregate}
\small
\tilde{Z}_v=\mathrm{diag}(\tilde{w})Z_v[\mathcal{I}_t],
\end{equation}
where $\mathrm{diag}(\tilde{w})$ denotes a diagonal matrix whose main diagonal is formed by $\tilde{w}$.

\begin{table*}[t]
\centering
\scriptsize
\setlength{\tabcolsep}{3.5pt}
\renewcommand{\arraystretch}{1.3}
\resizebox{\textwidth}{!}{
\begin{tabular}{llcccccccc}
\toprule[1pt]
\multirow{2}{*}{\textbf{Models}} & \multirow{2}{*}{\textbf{Methods}} & \multicolumn{2}{c}{\textbf{Random}} & \multicolumn{2}{c}{\textbf{Popular}} & \multicolumn{2}{c}{\textbf{Adversarial}} & \multicolumn{2}{c}{\textbf{Average}} \\
\cmidrule(lr){3-4} \cmidrule(lr){5-6} \cmidrule(lr){7-8} \cmidrule(lr){9-10}
 & & \textbf{Acc (\%)$\uparrow$} & \textbf{F1 (\%)$\uparrow$} & \textbf{Acc (\%)$\uparrow$} & \textbf{F1 (\%)$\uparrow$} & \textbf{Acc (\%)$\uparrow$} & \textbf{F1 (\%)$\uparrow$} & \textbf{Acc (\%)$\uparrow$} & \textbf{F1 (\%)$\uparrow$} \\
\midrule
\multirow{6}{*}{\textbf{LLaVA-1.5}} & Vanilla & 83.8 {\color{gray}$\uparrow$0.0} & 84.2 {\color{gray}$\uparrow$0.0} & 82.0 {\color{gray}$\uparrow$0.0} & 82.6 {\color{gray}$\uparrow$0.0} & 75.8 {\color{gray}$\uparrow$0.0} & 78.1 {\color{gray}$\uparrow$0.0} & 80.5 {\color{gray}$\uparrow$0.0} & 81.6 {\color{gray}$\uparrow$0.0}\\
 & +VCD \citep{leng2024mitigating} & 85.0 {\color{green!50!black}$\uparrow$1.2} & 84.2 {\color{gray}$\uparrow$0.0} & 82.1 {\color{green!50!black}$\uparrow$0.1} & 83.2 {\color{green!50!black}$\uparrow$0.6} & 76.3 {\color{green!50!black}$\uparrow$0.5} & 78.7 {\color{green!50!black}$\uparrow$0.6} & 81.1 {\color{green!50!black}$\uparrow$0.6} & 82.0 {\color{green!50!black}$\uparrow$0.4}\\
 & +M3ID \citep{favero2024multi} & 86.1 {\color{green!50!black}$\uparrow$2.3} & 85.0 {\color{green!50!black}$\uparrow$0.8} & 82.8 {\color{green!50!black}$\uparrow$0.8} & 84.1 {\color{green!50!black}$\uparrow$1.5} & 77.1 {\color{green!50!black}$\uparrow$1.3} & 78.9 {\color{green!50!black}$\uparrow$0.8} & 82.0 {\color{green!50!black}$\uparrow$1.5} & 82.7 {\color{green!50!black}$\uparrow$1.1}\\
 & +VTI \citep{liu2025reducing} & 83.0 {\color{red}$\downarrow$0.8} & 83.6 {\color{red}$\downarrow$0.6} & 80.4 {\color{red}$\downarrow$1.6} & 81.8 {\color{red}$\downarrow$0.8} & 76.0 {\color{green!50!black}$\uparrow$0.2} & 78.8 {\color{green!50!black}$\uparrow$0.7} & 79.8 {\color{red}$\downarrow$0.7} & 81.4 {\color{red}$\downarrow$0.2}\\
 & +AvisC \citep{woo2025don} & 82.3 {\color{red}$\downarrow$1.5} & 83.5 {\color{red}$\downarrow$0.7} & 78.2 {\color{red}$\downarrow$3.8} & 80.5 {\color{red}$\downarrow$2.1} & 74.2 {\color{red}$\downarrow$1.6} & 77.7 {\color{red}$\downarrow$0.4} & 78.2 {\color{red}$\downarrow$2.3} & 80.6 {\color{red}$\downarrow$1.0}\\
 \rowcolor{blue!4}
 & +\textbf{RVSD (Ours)} & \textbf{87.3} {\color{green!50!black}$\uparrow$3.5} & \textbf{86.3} {\color{green!50!black}$\uparrow$2.1} & \textbf{86.6} {\color{green!50!black}$\uparrow$4.6} & \textbf{85.3} {\color{green!50!black}$\uparrow$2.7} & \textbf{84.7} {\color{green!50!black}$\uparrow$8.9} & \textbf{83.5} {\color{green!50!black}$\uparrow$5.4} & \textbf{86.2} {\color{green!50!black}$\uparrow$5.7} & \textbf{85.0} {\color{green!50!black}$\uparrow$3.4}\\
\midrule
\multirow{6}{*}{\textbf{LLaVA-NEXT}} & Vanilla & 84.4 {\color{gray}$\uparrow$0.0} & 82.3 {\color{gray}$\uparrow$0.0} & 83.2 {\color{gray}$\uparrow$0.0} & 81.5 {\color{gray}$\uparrow$0.0} & 79.5 {\color{gray}$\uparrow$0.0} & 78.1 {\color{gray}$\uparrow$0.0} & 82.4 {\color{gray}$\uparrow$0.0} & 80.6 {\color{gray}$\uparrow$0.0}\\
 & +VCD \citep{leng2024mitigating} & 86.0 {\color{green!50!black}$\uparrow$1.6} & 84.3 {\color{green!50!black}$\uparrow$2.0} & 84.5 {\color{green!50!black}$\uparrow$1.3} & 82.9 {\color{green!50!black}$\uparrow$1.4} & 80.9 {\color{green!50!black}$\uparrow$1.4} & 79.7 {\color{green!50!black}$\uparrow$1.6} & 83.8 {\color{green!50!black}$\uparrow$1.4} & 82.3 {\color{green!50!black}$\uparrow$1.7}\\
 & +M3ID \citep{favero2024multi} & 85.5 {\color{green!50!black}$\uparrow$1.1} & 83.6 {\color{green!50!black}$\uparrow$1.3} & 84.2 {\color{green!50!black}$\uparrow$1.0} & 82.4 {\color{green!50!black}$\uparrow$0.9} & 80.6 {\color{green!50!black}$\uparrow$1.1} & 79.2 {\color{green!50!black}$\uparrow$1.1} & 83.4 {\color{green!50!black}$\uparrow$1.0} & 81.7 {\color{green!50!black}$\uparrow$1.1}\\
 & +VTI \citep{liu2025reducing} & 84.8 {\color{green!50!black}$\uparrow$0.4} & 83.1 {\color{green!50!black}$\uparrow$0.8} & 81.8 {\color{red}$\downarrow$1.4} & 81.6 {\color{green!50!black}$\uparrow$0.1} & 79.0 {\color{red}$\downarrow$0.5} & 78.1 {\color{gray}$\uparrow$0.0} & 81.9 {\color{red}$\downarrow$0.5} & 80.9 {\color{green!50!black}$\uparrow$0.3}\\
 & +AvisC \citep{woo2025don} & 85.2 {\color{green!50!black}$\uparrow$0.8} & 82.8 {\color{green!50!black}$\uparrow$0.5} & 83.9 {\color{green!50!black}$\uparrow$0.7} & 81.6 {\color{green!50!black}$\uparrow$0.1} & 81.8 {\color{green!50!black}$\uparrow$2.3} & 80.2 {\color{green!50!black}$\uparrow$2.1} & 83.6 {\color{green!50!black}$\uparrow$1.2} & 81.5 {\color{green!50!black}$\uparrow$0.9}\\
 \rowcolor{blue!4}
 & +\textbf{RVSD (Ours)} & \textbf{88.5} {\color{green!50!black}$\uparrow$4.1} & \textbf{87.6} {\color{green!50!black}$\uparrow$5.3} & \textbf{87.9} {\color{green!50!black}$\uparrow$4.7} & \textbf{86.7} {\color{green!50!black}$\uparrow$5.2} & \textbf{86.5} {\color{green!50!black}$\uparrow$7.0} & \textbf{85.4} {\color{green!50!black}$\uparrow$7.3} & \textbf{87.6} {\color{green!50!black}$\uparrow$5.2} & \textbf{86.6} {\color{green!50!black}$\uparrow$6.0}\\
\midrule
\multirow{6}{*}{\textbf{Qwen-VL}} & Vanilla & 84.9 {\color{gray}$\uparrow$0.0} & 82.9 {\color{gray}$\uparrow$0.0} & 84.0 {\color{gray}$\uparrow$0.0} & 81.9 {\color{gray}$\uparrow$0.0} & 82.1 {\color{gray}$\uparrow$0.0} & 80.2 {\color{gray}$\uparrow$0.0} & 83.7 {\color{gray}$\uparrow$0.0} & 81.7 {\color{gray}$\uparrow$0.0}\\
 & +VCD \citep{leng2024mitigating} & 85.5 {\color{green!50!black}$\uparrow$0.6} & 83.6 {\color{green!50!black}$\uparrow$0.7} & 84.9 {\color{green!50!black}$\uparrow$0.9} & \textbf{83.6} {\color{green!50!black}$\uparrow$1.7} & 84.0 {\color{green!50!black}$\uparrow$1.9} & 82.0 {\color{green!50!black}$\uparrow$1.8} & 84.8 {\color{green!50!black}$\uparrow$1.1} & 83.1 {\color{green!50!black}$\uparrow$1.4}\\
 & +M3ID \citep{favero2024multi} & 85.3 {\color{green!50!black}$\uparrow$0.4} & 83.4 {\color{green!50!black}$\uparrow$0.5} & 84.2 {\color{green!50!black}$\uparrow$0.2} & 82.7 {\color{green!50!black}$\uparrow$0.8} & 83.2 {\color{green!50!black}$\uparrow$1.1} & 80.8 {\color{green!50!black}$\uparrow$0.6} & 84.2 {\color{green!50!black}$\uparrow$0.5} & 82.3 {\color{green!50!black}$\uparrow$0.6}\\
 & +VTI \citep{liu2025reducing} & 85.3 {\color{green!50!black}$\uparrow$0.4} & 83.5 {\color{green!50!black}$\uparrow$0.6} & 83.0 {\color{red}$\downarrow$1.0} & 82.3 {\color{green!50!black}$\uparrow$0.4} & 83.2 {\color{green!50!black}$\uparrow$1.1} & 81.8 {\color{green!50!black}$\uparrow$1.6} & 83.8 {\color{green!50!black}$\uparrow$0.1} & 82.5 {\color{green!50!black}$\uparrow$0.8}\\
 & +AvisC \citep{woo2025don} & 82.9 {\color{red}$\downarrow$2.0} & 80.0 {\color{red}$\downarrow$2.9} & 82.8 {\color{red}$\downarrow$1.2} & 80.1 {\color{red}$\downarrow$1.8} & 81.2 {\color{red}$\downarrow$0.9} & 78.5 {\color{red}$\downarrow$1.7} & 82.3 {\color{red}$\downarrow$1.4} & 79.5 {\color{red}$\downarrow$2.2}\\
 \rowcolor{blue!4}
 & +\textbf{RVSD (Ours)} & \textbf{86.2} {\color{green!50!black}$\uparrow$1.3} & \textbf{84.7} {\color{green!50!black}$\uparrow$1.8} & \textbf{85.3} {\color{green!50!black}$\uparrow$1.3} & 83.3 {\color{green!50!black}$\uparrow$1.4} & \textbf{84.5} {\color{green!50!black}$\uparrow$2.4} & \textbf{82.6} {\color{green!50!black}$\uparrow$2.4} & \textbf{85.3} {\color{green!50!black}$\uparrow$1.6} & \textbf{83.5} {\color{green!50!black}$\uparrow$1.8}\\
\bottomrule[1pt]
\end{tabular}
}
\captionsetup{font=small}
\caption{\label{tab:chair_pope_compare} Performance comparison of RVSD with the baselines on the POPE-MSCOCO discrimination benchmark across the random, popular, and adversarial splits. The average column reports the mean accuracy and F1 scores over all splits. Green ({\color{green!50!black}$\bullet$}) and red ({\color{red}$\bullet$}) indicate improvements and degradations relative to the base model, respectively, while ($\uparrow$/$\downarrow$) denotes the absolute change. Best results are highlighted in \textbf{bold}.}
\end{table*}

\subsubsection{Lightweight injection}
To inject the retrieved evidence into the decoding stream without disturbing the model’s intrinsic distribution, we introduce a transient adapter whose weights are scale-matched to the host Feed-Forward Network (FFN), expressed as
\begin{equation}\label{eq:adapter_w}
\small
W_1^{\mathrm{adpt}} = \frac{s_{\mathrm{up}}}{\tilde{s}_v}\tilde{Z}_v, \:\:
W_2^{\mathrm{adpt}} = \frac{s_{\mathrm{down}}}{\tilde{s}_v}\tilde{Z}_v^{\top}.
\end{equation}
Here, $W_1^{\mathrm{adpt}}$ and $W_2^{\mathrm{adpt}}$ denote the up- and down-projection weights of the transient adapter, respectively. $W_{\mathrm{up}}$ and $W_{\mathrm{down}}$ denote the corresponding up- and down-projection matrices of the host FFN, respectively. $s_{\mathrm{up}}=\mathrm{mean}(|W_{\mathrm{up}}|)$, $s_{\mathrm{down}}=\mathrm{mean}(|W_{\mathrm{down}}|)$, and $\tilde{s}_v=\mathrm{mean}(|\tilde{Z}_v|)$ denote the mean absolute magnitudes used for amplitude alignment. The adapter performs a low-rank projection that re-grounds the textual representation using the retrieved visual evidence, expressed as
\begin{equation}
\small
\mathcal{G}(x) = x(W_1^{\mathrm{adpt}})^{\top}(W_2^{\mathrm{adpt}})^{\top} = \frac{s_{\mathrm{up}}s_{\mathrm{down}}x\tilde{Z}_v^{\top}\tilde{Z}_v}{\tilde{s}_v^2},
\end{equation}
where $x$ denotes the input residual-stream hidden state fed into the adapter. After amplitude normalization, the retrieved visual signal $\hat{a}$ is fused with the block-output residual stream $h_t^{(l+1)}$, i.e., the hidden state after FFN residual addition in layer $l$ that feeds into the attention sublayer of layer $l{+}1$:
\begin{equation}\label{eq:fusion}
\small
o=(1-\alpha)\cdot h_t^{(l+1)}+\alpha\cdot\hat{a},
\end{equation}
where $o$ denotes the fused hidden state passed to layer $l{+}1$, and $\alpha \in [0,1]$ denotes the injection ratio that balances the original residual stream and the retrieved visual evidence. After retrieval, the trigger is reset, and the transient adapter is released. This enables parameter-free and on-demand visual retrieval in the semantic space, i.e., visual evidence is retrieved only when grounding fails, and the overhead is removed after each step, ensuring efficiency while mitigating VHs.

\section{Experiments}
\subsection{Experimental Settings}

\noindent\textbf{Benchmarks.}
We evaluate RVSD on five complementary benchmarks. For discriminative hallucination, we use POPE~\citep{li2023evaluating} with random, popular, and adversarial splits, the MME hallucination subset~\citep{fu2026mme} covering object-level and attribute-level probes, and the discrimination subset of AMBER~\citep{wang2023amber}, reporting accuracy and F1 scores. For generative hallucination, we evaluate on CHAIR~\citep{rohrbach2018object} over 500 MS-COCO images and the AMBER generative subset, and also report AMBER discrimination accuracy and F1 scores for a unified view of caption-level performance. To assess whether VH mitigation affects general capability, we further evaluate on MM-Vet~\citep{pmlr-v235-yu24o}, which covers six vision-language abilities and uses GPT-4 as the evaluator.

\smallskip\noindent\textbf{Models.}
We evaluate RVSD on the 7B variants of LLaVA-1.5~\citep{liu2024improved}, LLaVA-NEXT~\citep{liu2024llavanext}, and Qwen-VL~\citep{bai2023qwen}. LLaVA-1.5 adopts a two-layer MLP for vision-language alignment, and LLaVA-NEXT further introduces the AnyRes mechanism to support higher-resolution visual understanding, while Qwen-VL instead couples a cross-attention-based vision-language adapter with a non-LLaMA language backbone. Together, they cover both standard- and high-resolution settings as well as heterogeneous vision-language interfaces, enabling a comprehensive evaluation of RVSD. Moreover, RVSD is model-agnostic and can be seamlessly integrated into different LVLM architectures.

\smallskip\noindent\textbf{Baselines.}
We compare RVSD with representative training-free baselines from three categories: (i) contrastive decoders, including VCD~\citep{leng2024mitigating} and M3ID~\citep{favero2024multi}; (ii) the attention-intervention method AvisC~\citep{woo2025don}; and (iii) the latent-space steering method VTI~\citep{liu2025reducing}. All baselines leverage the hyperparameters recommended in their original papers under the same decoding budget, while the vanilla LVLM with sampling decoding serves as the base model.

\begin{figure}[t]
\centering
\includegraphics[width=0.45\textwidth]{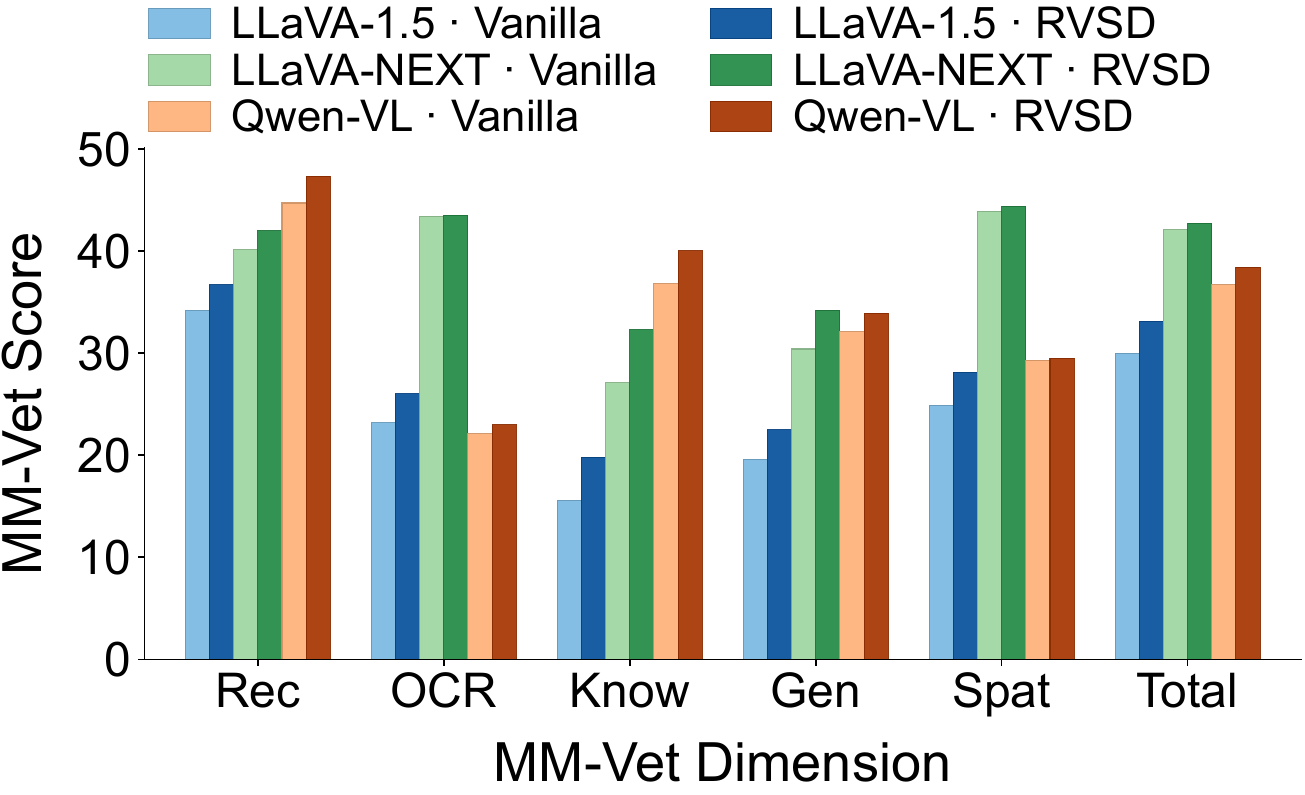}
\captionsetup{font=small}
\caption{\label{fig:mmvet} Per-dimension MM-Vet scores of Vanilla decoding and RVSD on LLaVA-1.5, LLaVA-NEXT, and Qwen-VL, covering the Rec, OCR, Know, Gen, and Spat capability dimensions as well as the overall total score. RVSD improves every reported dimension on all three backbones.}
\end{figure}

\begin{figure*}[t]
    \centering
    \includegraphics[width=\textwidth]{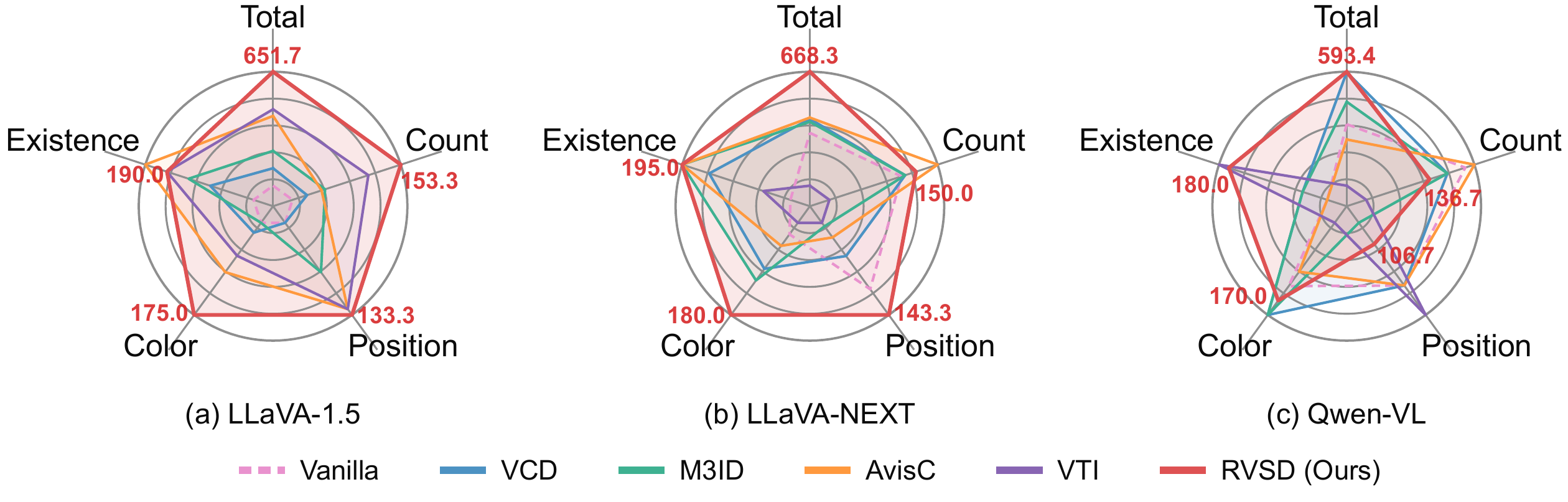}
    \captionsetup{font=small}
    \caption{\label{fig:mme} Performance comparison of RVSD with the baselines on the MME hallucination subset for (a) LLaVA-1.5, (b) LLaVA-NEXT, and (c) Qwen-VL, covering object-level (Existence, Count) and attribute-level (Position, Color) probes, with Total denoting their sum. Values annotated in {\color{red}red} are those of RVSD, and each axis is scaled independently.}
\end{figure*}

\subsection{Experimental Results}

\noindent\textbf{Results on POPE.}
In Table~\ref{tab:chair_pope_compare}, we see that RVSD achieves the best performance across all POPE splits on the three backbones, obtaining the highest accuracy on every split and the highest F1 scores in almost all cases. On LLaVA-1.5-7B, RVSD improves the average Accuracy/F1 from 80.5/81.6 to 86.2/85.0. On LLaVA-NEXT, the scores increase from 82.4/80.6 to 87.6/86.6, and on Qwen-VL, from 83.7/81.7 to 85.3/83.5. The largest gains are observed on the adversarial split, with improvements of +8.9/+5.4, +7.0/+7.3, and +2.4/+2.4 on the three backbones, respectively. In contrast, VCD and M3ID yield only limited improvements, while VTI and AvisC occasionally degrade performance. These results demonstrate the effectiveness of combining semantics-directed sparsification with on-demand retrieval for reliable visual grounding.

\smallskip\noindent\textbf{Results on MM-Vet.}
To verify that VH mitigation does not compromise general multimodal capability, we evaluate RVSD on MM-Vet and report the Rec, OCR, Know, Gen, and Spat dimensions together with the overall total score in Fig.~\ref{fig:mmvet}. We observe that RVSD improves the total score over the vanilla model on all three backbones, from 30.0 to 33.1 on LLaVA-1.5, from 42.1 to 42.7 on LLaVA-NEXT, and from 36.7 to 38.4 on Qwen-VL, with gains observed across every reported dimension. The improvements are most pronounced on Know, suggesting that restoring pruned visual evidence also benefits knowledge-intensive reasoning. The results demonstrate that RVSD mitigates VHs while preserving general multimodal competence.

\smallskip\noindent\textbf{Results on MME.}
As shown in Fig.~\ref{fig:mme}, RVSD obtains the highest total score on the MME hallucination subset across all three backbones, reaching 651.7 on LLaVA-1.5, 668.3 on LLaVA-NEXT, and 593.4 on Qwen-VL, and consistently surpassing the vanilla decoder on the total axis of every panel. On the two LLaVA backbones, the improvements are concentrated on the attribute-level probes, where RVSD achieves the highest scores on both Color and Position, whereas the scores of most baselines decline markedly on Position for LLaVA-NEXT. On Qwen-VL, the gain is instead driven by object Existence, while Count and Position remain below the strongest baselines. These results further demonstrate the effectiveness of RVSD in mitigating fine-grained VHs.

\begin{table*}[t]
\centering
\scriptsize
\setlength{\tabcolsep}{3.5pt}
\renewcommand{\arraystretch}{1.3}
\resizebox{\textwidth}{!}{
\begin{tabular}{llccccccccc}
\toprule[1pt]
\multirow{2}{*}{\textbf{Models}} & \multirow{2}{*}{\textbf{Methods}} & \multicolumn{4}{c}{\textbf{CHAIR}} & \multicolumn{5}{c}{\textbf{AMBER}} \\
\cmidrule(lr){3-6} \cmidrule(lr){7-11}
 & & \textbf{CHAIR$_S$$\downarrow$} & \textbf{CHAIR$_I$$\downarrow$} & \textbf{Recall$\uparrow$} & \textbf{Length} & \textbf{Acc (\%)$\uparrow$} & \textbf{F1 (\%)$\uparrow$} & \textbf{CHAIR$\downarrow$} & \textbf{HalRate$\downarrow$} & \textbf{Score$\uparrow$} \\
\midrule
\multirow{6}{*}{\textbf{LLaVA-1.5}} & Vanilla & 52.8 {\color{gray}$\downarrow$0.0} & 15.9 {\color{gray}$\downarrow$0.0} & 77.3 {\color{gray}$\uparrow$0.0} & \phantom{0}93.4 & 67.0 {\color{gray}$\uparrow$0.0} & 71.0 {\color{gray}$\uparrow$0.0} & 12.0 {\color{gray}$\downarrow$0.0} & 51.0 {\color{gray}$\downarrow$0.0} & 79.5 {\color{gray}$\uparrow$0.0}\\
 & +VCD \citep{leng2024mitigating} & 51.0 {\color{green!50!black}$\downarrow$1.8} & 14.9 {\color{green!50!black}$\downarrow$1.0} & 77.2 {\color{red}$\downarrow$0.1} & 101.9 & 67.3 {\color{green!50!black}$\uparrow$0.3} & 71.1 {\color{green!50!black}$\uparrow$0.1} & 10.0 {\color{green!50!black}$\downarrow$2.0} & 43.6 {\color{green!50!black}$\downarrow$7.4} & 80.6 {\color{green!50!black}$\uparrow$1.1}\\
 & +M3ID \citep{favero2024multi} & 56.2 {\color{red}$\uparrow$3.4} & 17.0 {\color{red}$\uparrow$1.1} & \textbf{79.3} {\color{green!50!black}$\uparrow$2.0} & \phantom{0}97.1 & 67.3 {\color{green!50!black}$\uparrow$0.3} & 70.9 {\color{red}$\downarrow$0.1} & \phantom{0}9.1 {\color{green!50!black}$\downarrow$2.9} & 42.8 {\color{green!50!black}$\downarrow$8.2} & 80.9 {\color{green!50!black}$\uparrow$1.4}\\
 & +AvisC \citep{woo2025don} & 44.0 {\color{green!50!black}$\downarrow$8.8} & 13.7 {\color{green!50!black}$\downarrow$2.2} & 72.9 {\color{red}$\downarrow$4.4} & \phantom{0}89.8 & 70.7 {\color{green!50!black}$\uparrow$3.7} & 74.2 {\color{green!50!black}$\uparrow$3.2} & 11.8 {\color{green!50!black}$\downarrow$0.2} & 51.2 {\color{red}$\uparrow$0.2} & 81.2 {\color{green!50!black}$\uparrow$1.7}\\
 & +VTI \citep{liu2025reducing} & \textbf{36.8} {\color{green!50!black}$\downarrow$16.} & 13.6 {\color{green!50!black}$\downarrow$2.3} & 66.6 {\color{red}$\downarrow$10.} & \phantom{0}66.7 & 66.5 {\color{red}$\downarrow$0.5} & 70.6 {\color{red}$\downarrow$0.4} & \phantom{0}\textbf{6.9} {\color{green!50!black}$\downarrow$5.1} & \textbf{27.0} {\color{green!50!black}$\downarrow$24.} & 81.9 {\color{green!50!black}$\uparrow$2.4}\\
 \rowcolor{blue!4}
 & +\textbf{RVSD (Ours)} & 38.4 {\color{green!50!black}$\downarrow$14.} & \textbf{10.3} {\color{green!50!black}$\downarrow$5.6} & 74.4 {\color{red}$\downarrow$2.9} & \phantom{0}90.6 & \textbf{73.6} {\color{green!50!black}$\uparrow$6.6} & \textbf{76.9} {\color{green!50!black}$\uparrow$5.9} & \phantom{0}7.3 {\color{green!50!black}$\downarrow$4.7} & 31.2 {\color{green!50!black}$\downarrow$19.} & \textbf{84.8} {\color{green!50!black}$\uparrow$5.3}\\
\midrule
\multirow{6}{*}{\textbf{LLaVA-NEXT}} & Vanilla & 35.8 {\color{gray}$\downarrow$0.0} & 12.0 {\color{gray}$\downarrow$0.0} & 59.5 {\color{gray}$\uparrow$0.0} & 179.0 & 72.9 {\color{gray}$\uparrow$0.0} & 78.6 {\color{gray}$\uparrow$0.0} & 12.0 {\color{gray}$\downarrow$0.0} & 59.6 {\color{gray}$\downarrow$0.0} & 83.3 {\color{gray}$\uparrow$0.0}\\
 & +VCD \citep{leng2024mitigating} & 40.2 {\color{red}$\uparrow$4.4} & 10.7 {\color{green!50!black}$\downarrow$1.3} & 62.1 {\color{green!50!black}$\uparrow$2.6} & 171.2 & 74.3 {\color{green!50!black}$\uparrow$1.4} & 79.6 {\color{green!50!black}$\uparrow$1.0} & 11.8 {\color{green!50!black}$\downarrow$0.2} & 58.6 {\color{green!50!black}$\downarrow$1.0} & 83.9 {\color{green!50!black}$\uparrow$0.6}\\
 & +M3ID \citep{favero2024multi} & 35.2 {\color{green!50!black}$\downarrow$0.6} & 10.3 {\color{green!50!black}$\downarrow$1.7} & 61.4 {\color{green!50!black}$\uparrow$1.9} & 152.7 & 75.0 {\color{green!50!black}$\uparrow$2.1} & 80.0 {\color{green!50!black}$\uparrow$1.4} & 10.2 {\color{green!50!black}$\downarrow$1.8} & 52.8 {\color{green!50!black}$\downarrow$6.8} & 84.9 {\color{green!50!black}$\uparrow$1.6}\\
 & +AvisC \citep{woo2025don} & 40.4 {\color{red}$\uparrow$4.6} & 12.4 {\color{red}$\uparrow$0.4} & 60.0 {\color{green!50!black}$\uparrow$0.5} & 183.8 & 77.7 {\color{green!50!black}$\uparrow$4.8} & 82.1 {\color{green!50!black}$\uparrow$3.5} & 14.2 {\color{red}$\uparrow$2.2} & 65.2 {\color{red}$\uparrow$5.6} & 84.0 {\color{green!50!black}$\uparrow$0.7}\\
 & +VTI \citep{liu2025reducing} & 26.4 {\color{green!50!black}$\downarrow$9.4} & \phantom{0}7.7 {\color{green!50!black}$\downarrow$4.3} & 58.5 {\color{red}$\downarrow$1.0} & 187.6 & 75.2 {\color{green!50!black}$\uparrow$2.3} & 81.9 {\color{green!50!black}$\uparrow$3.3} & 10.3 {\color{green!50!black}$\downarrow$1.7} & 58.3 {\color{green!50!black}$\downarrow$1.3} & 87.1 {\color{green!50!black}$\uparrow$3.8}\\
 \rowcolor{blue!4}
 & +\textbf{RVSD (Ours)} & \textbf{26.2} {\color{green!50!black}$\downarrow$9.6} & \phantom{0}\textbf{6.7} {\color{green!50!black}$\downarrow$5.3} & \textbf{63.7} {\color{green!50!black}$\uparrow$4.2} & 185.0 & \textbf{80.0} {\color{green!50!black}$\uparrow$7.1} & \textbf{83.8} {\color{green!50!black}$\uparrow$5.2} & \phantom{0}\textbf{8.9} {\color{green!50!black}$\downarrow$3.1} & \textbf{47.9} {\color{green!50!black}$\downarrow$11.} & \textbf{87.5} {\color{green!50!black}$\uparrow$4.2}\\
\midrule
\multirow{6}{*}{\textbf{Qwen-VL}} & Vanilla & \phantom{0}2.8 {\color{gray}$\downarrow$0.0} & \phantom{0}3.0 {\color{gray}$\downarrow$0.0} & 31.0 {\color{gray}$\uparrow$0.0} & \phantom{00}5.3 & 82.9 {\color{gray}$\uparrow$0.0} & 86.9 {\color{gray}$\uparrow$0.0} & \phantom{0}4.8 {\color{gray}$\downarrow$0.0} & \phantom{0}9.2 {\color{gray}$\downarrow$0.0} & 91.1 {\color{gray}$\uparrow$0.0}\\
 & +VCD \citep{leng2024mitigating} & \phantom{0}1.4 {\color{green!50!black}$\downarrow$1.4} & \phantom{0}\textbf{1.2} {\color{green!50!black}$\downarrow$1.8} & 30.8 {\color{red}$\downarrow$0.2} & \phantom{00}4.0 & 84.1 {\color{green!50!black}$\uparrow$1.2} & 87.9 {\color{green!50!black}$\uparrow$1.0} & \phantom{0}3.5 {\color{green!50!black}$\downarrow$1.3} & \phantom{0}7.7 {\color{green!50!black}$\downarrow$1.5} & 92.2 {\color{green!50!black}$\uparrow$1.1}\\
 & +M3ID \citep{favero2024multi} & \phantom{0}1.7 {\color{green!50!black}$\downarrow$1.1} & \phantom{0}1.3 {\color{green!50!black}$\downarrow$1.7} & 31.8 {\color{green!50!black}$\uparrow$0.8} & \phantom{00}3.4 & 84.0 {\color{green!50!black}$\uparrow$1.1} & 87.1 {\color{green!50!black}$\uparrow$0.2} & \phantom{0}3.7 {\color{green!50!black}$\downarrow$1.1} & \phantom{0}8.2 {\color{green!50!black}$\downarrow$1.0} & 91.7 {\color{green!50!black}$\uparrow$0.6}\\
 & +AvisC \citep{woo2025don} & \phantom{0}1.6 {\color{green!50!black}$\downarrow$1.2} & \phantom{0}1.6 {\color{green!50!black}$\downarrow$1.4} & \textbf{32.0} {\color{green!50!black}$\uparrow$1.0} & \phantom{00}4.4 & 84.1 {\color{green!50!black}$\uparrow$1.2} & 87.9 {\color{green!50!black}$\uparrow$1.0} & \phantom{0}4.3 {\color{green!50!black}$\downarrow$0.5} & \phantom{0}8.6 {\color{green!50!black}$\downarrow$0.6} & 91.8 {\color{green!50!black}$\uparrow$0.7}\\
 & +VTI \citep{liu2025reducing} & \phantom{0}1.9 {\color{green!50!black}$\downarrow$0.9} & \phantom{0}1.9 {\color{green!50!black}$\downarrow$1.1} & 30.2 {\color{red}$\downarrow$0.8} & \phantom{00}4.5 & 83.5 {\color{green!50!black}$\uparrow$0.6} & 87.3 {\color{green!50!black}$\uparrow$0.4} & \phantom{0}\textbf{3.1} {\color{green!50!black}$\downarrow$1.7} & \phantom{0}\textbf{6.8} {\color{green!50!black}$\downarrow$2.4} & 92.4 {\color{green!50!black}$\uparrow$1.3}\\
 \rowcolor{blue!4}
 & +\textbf{RVSD (Ours)} & \phantom{0}\textbf{1.2} {\color{green!50!black}$\downarrow$1.6} & \phantom{0}1.6 {\color{green!50!black}$\downarrow$1.4} & 31.9 {\color{green!50!black}$\uparrow$0.9} & \phantom{00}5.9 & \textbf{84.3} {\color{green!50!black}$\uparrow$1.4} & \textbf{88.6} {\color{green!50!black}$\uparrow$1.7} & \phantom{0}3.4 {\color{green!50!black}$\downarrow$1.4} & \phantom{0}7.3 {\color{green!50!black}$\downarrow$1.9} & \textbf{92.6} {\color{green!50!black}$\uparrow$1.5}\\
\bottomrule[1pt]
\end{tabular}
}
\captionsetup{font=small}
\caption{\label{tab:amber_results}Performance comparison on the CHAIR benchmark (CHAIR$_S$, CHAIR$_I$, Recall, Length) and the AMBER benchmark for open-ended captioning, where the AMBER block jointly reports discrimination metrics (Accuracy, F1) and generative metrics (CHAIR, HalRate, Score). Best results are highlighted in \textbf{bold}.}
\end{table*}

\smallskip\noindent\textbf{Results on CHAIR and AMBER.}
As shown in Table~\ref{tab:amber_results}, we further evaluate caption-level behavior on CHAIR and AMBER, jointly assessing discrimination (Accuracy/F1) and generative hallucination (CHAIR$_S$, CHAIR$_I$, AMBER CHAIR, HalRate, Score). We observe that RVSD achieves the best discrimination performance and the highest AMBER score across all three backbones, and obtains the lowest CHAIR$_I$ on the two LLaVA backbones and the lowest CHAIR$_S$ on Qwen-VL, indicating effective suppression of object-level hallucinations during long-form generation. In contrast, existing baselines provide only limited or inconsistent improvements. Overall, RVSD consistently improves both faithfulness and caption quality across different LVLM architectures, demonstrating the effectiveness of combining semantics-directed sparsification with on-demand visual retrieval.

\subsection{Ablation Studies}

\noindent\textbf{Mechanism Contribution.}
To evaluate the contribution of each component, we conduct an ablation study on the full MME benchmark using LLaVA-1.5 by separately removing the semantics-directed sparse selection (w/o Sparse) and the SSVR module (w/o SSVR). As shown in Table~\ref{tab:component_ablation}, removing either component degrades the overall total score, demonstrating that both modules are essential to RVSD. In particular, removing SSVR causes the largest performance drop, reducing Cognition from 371.79 to 348.21 and yielding the greatest decline in the total score, highlighting the importance of on-demand visual retrieval for restoring visual grounding during reasoning-intensive decoding. In contrast, removing sparse selection slightly improves the perception score by retaining all visual tokens, but noticeably reduces the cognition score by 14.29, indicating that dense decoding introduces redundant visual information that weakens effective grounding. Overall, RVSD achieves the best total and cognition scores, demonstrating that semantics-directed sparsification and SSVR complement each other to improve visual grounding.

\begin{table}[t]
\centering
\tiny
\setlength{\tabcolsep}{2.8pt}
\renewcommand{\arraystretch}{1.3}
\resizebox{0.95\columnwidth}{!}{
\begin{tabular}{lccc}
\toprule[0.6pt]
\rowcolor{gray!8}
\textbf{Methods} & \textbf{Perception}$\uparrow$ & \textbf{Cognition}$\uparrow$ & \textbf{Total}$\uparrow$ \\
\midrule
RVSD        & 1511.40 & \textbf{371.79} & \textbf{1883.19} \\
w/o Sparse  & 1514.60 & 357.50 & 1872.10 \\
w/o SSVR    & \textbf{1520.36} & 348.21 & 1868.57 \\
\bottomrule[0.6pt]
\end{tabular}
}
\captionsetup{font=small}
\caption{\label{tab:component_ablation} Component-level ablation study on the full MME benchmark using LLaVA-1.5, evaluating the contributions of the semantics-directed sparse selection module and the SSVR module. Best results are highlighted in \textbf{bold}.}
\end{table}

\subsection{Inference Latency}

\begin{table}[t]
\centering
\scriptsize
\setlength{\tabcolsep}{3pt}
\renewcommand{\arraystretch}{1.3}
\resizebox{0.95\columnwidth}{!}{
\begin{tabular}{lcccc}
\toprule[0.8pt]
\rowcolor{gray!8}
\textbf{Methods} & \textbf{Latency} $\downarrow$ & \textbf{Memory} $\downarrow$ & \textbf{TFLOPs} $\downarrow$ & \textbf{TTFT} $\downarrow$ \\
\midrule
Vanilla & \phantom{0}\textbf{61.9} & 14334 & 4.36 & 243.8 \\
+VCD & 131.6 & 14979 & 8.72 & 445.4 \\
+M3ID & 132.5 & 14977 & 8.72 & 477.8 \\
+AvisC & 173.6 & 15501 & 8.45 & 725.2 \\
+VTI & \phantom{0}67.3 & 14363 & 4.36 & 266.3 \\
+RVSD & \phantom{0}62.0 & \textbf{14015} & \textbf{3.68} & \textbf{194.6} \\
\bottomrule[0.8pt]
\end{tabular}
}
\captionsetup{font=small}
\caption{\label{tab:latency} Inference efficiency comparison between RVSD and representative decoding-side VH mitigation methods, reporting per-token decoding latency ($\mathrm{ms/token}$), peak GPU memory usage ($\mathrm{MB}$), TFLOPs ($\mathrm{T}$), and TTFT ($\mathrm{ms}$). Best results are highlighted in \textbf{bold}.}
\end{table}

As shown in Table~\ref{tab:latency}, we further benchmark inference efficiency under the same decoding budget. We observe that RVSD is the only method that maintains the same latency as the Vanilla baseline (1.00$\times$) while simultaneously reducing memory usage by 2.2\%, Tera Floating-Point Operations Per Second (TFLOPs) by 15.6\%, and Time-to-First-Token (TTFT) by 20.2\%. In contrast, contrastive decoding methods (VCD and M3ID) introduce over 2$\times$ latency overhead due to auxiliary forward passes, while AvisC incurs the highest cost from dual forward inference and additional attention extraction. Although VTI is relatively lightweight, it still increases latency by 1.09$\times$. These results demonstrate that RVSD effectively improves visual grounding without sacrificing inference efficiency, benefiting from semantics-directed sparsification and the uncertainty-triggered retrieval mechanism that is activated only when necessary.

\section{Conclusion}

In this paper, we have proposed RVSD, a training-free and plug-and-play decoding framework that, for the first time, resolves the previously overlooked sparsification-hallucination paradox in existing sparse decoders. By integrating semantics-directed token selection with the SSVR mechanism within a single decoding pass, RVSD transforms irreversible visual token pruning into reversible and on-demand cross-modal retrieval, enabling visual grounding to be restored only when necessary. Extensive experiments on five representative benchmarks across three LVLM backbones demonstrate that RVSD achieves SOTA performance in VH mitigation while simultaneously improving decoding efficiency and maintaining robustness under long-context generation, where existing sparse and decoding-based methods often deteriorate.

For future work, we will extend RVSD to more challenging multimodal settings, such as long-form video understanding, where extended temporal dependencies and richer modality interactions further exacerbate the sparsification-hallucination paradox. We hope this work provides a promising direction toward trustworthy and efficient LVLM inference.

\section*{Limitations}

While RVSD shows promising results, several limitations remain. Firstly, the uncertainty-triggered retrieval gate in SSVR relies on a predefined entropy threshold to determine when cross-modal retrieval should be activated. Although RVSD remains robust across a reasonable range of threshold values, a more adaptive gating strategy that dynamically adjusts the trigger according to input complexity can further reduce unnecessary retrieval calls and improve efficiency on simpler queries.

Additionally, the current implementation uses a simple dot-product similarity to retrieve evidence from the deferred visual memory bank. More advanced retrieval mechanisms, such as learned semantic projections or multi-head cross-modal attention, may further improve performance on fine-grained tasks. Although our evaluation already spans two distinct architectural families, i.e., the LLaVA series and Qwen-VL, we also plan to extend RVSD to larger-scale and more heterogeneous LVLMs to further evaluate its generality.

\section*{Ethical Considerations}

This work aims to improve the trustworthiness of LVLMs by mitigating VHs, thereby contributing to the safer deployment of multimodal artificial intelligence systems. RVSD is a training-free and inference-only framework that requires no additional data collection or human annotation. All datasets (i.e., POPE, MME, AMBER, CHAIR, MM-Vet, MSCOCO, A-OKVQA, and GQA) and backbone models (LLaVA-1.5, LLaVA-NEXT, and Qwen-VL) used in this study are publicly available for academic research and are used strictly in accordance with their original licenses. While RVSD substantially reduces VHs, it does not eliminate them. Therefore, we recommend pairing it with downstream verification or human oversight before deployment in safety-critical applications such as medical diagnosis or assistive technologies.

\section*{Acknowledgements}

This work was supported in part by the National Natural Science Foundation of China (NSFC) under Grant No. 62572132, and the Guangdong Basic and Applied Basic Research Foundation under Grant No. 2025A1515010137.

\bibliography{custom}

\clearpage

\appendix

\section*{Appendix}

\begin{table*}[t]
\centering
\scriptsize
\setlength{\tabcolsep}{3.5pt}
\renewcommand{\arraystretch}{1.3}
\resizebox{\textwidth}{!}{
\begin{tabular}{llcccccccc}
\toprule[1pt]
\multirow{2}{*}{\textbf{Models}} & \multirow{2}{*}{\textbf{Methods}} & \multicolumn{2}{c}{\textbf{Random}} & \multicolumn{2}{c}{\textbf{Popular}} & \multicolumn{2}{c}{\textbf{Adversarial}} & \multicolumn{2}{c}{\textbf{Average}} \\
\cmidrule(lr){3-4} \cmidrule(lr){5-6} \cmidrule(lr){7-8} \cmidrule(lr){9-10}
& & \textbf{Acc (\%)$\uparrow$} & \textbf{F1 (\%)$\uparrow$} & \textbf{Acc (\%)$\uparrow$} & \textbf{F1 (\%)$\uparrow$} & \textbf{Acc (\%)$\uparrow$} & \textbf{F1 (\%)$\uparrow$} & \textbf{Acc (\%)$\uparrow$} & \textbf{F1 (\%)$\uparrow$}\\
\midrule
\multirow{6}{*}{\textbf{LLaVA-1.5}} & Vanilla & 81.8 {\color{gray}$\uparrow$0.0} & 83.5 {\color{gray}$\uparrow$0.0} & 75.3 {\color{gray}$\uparrow$0.0} & 78.7 {\color{gray}$\uparrow$0.0} & 67.4 {\color{gray}$\uparrow$0.0} & 73.7 {\color{gray}$\uparrow$0.0} & 74.8 {\color{gray}$\uparrow$0.0} & 78.6 {\color{gray}$\uparrow$0.0}\\
 & +VCD \citep{leng2024mitigating} & 81.2 {\color{red}$\downarrow$0.6} & 83.2 {\color{red}$\downarrow$0.3} & 74.7 {\color{red}$\downarrow$0.6} & 78.5 {\color{red}$\downarrow$0.2} & 68.1 {\color{green!50!black}$\uparrow$0.7} & 74.6 {\color{green!50!black}$\uparrow$0.9} & 74.7 {\color{red}$\downarrow$0.1} & 78.8 {\color{green!50!black}$\uparrow$0.2}\\
 & +M3ID \citep{favero2024multi} & 82.9 {\color{green!50!black}$\uparrow$1.1} & 84.6 {\color{green!50!black}$\uparrow$1.1} & 75.8 {\color{green!50!black}$\uparrow$0.5} & 79.4 {\color{green!50!black}$\uparrow$0.7} & 68.3 {\color{green!50!black}$\uparrow$0.9} & 74.6 {\color{green!50!black}$\uparrow$0.9} & 75.7 {\color{green!50!black}$\uparrow$0.9} & 79.5 {\color{green!50!black}$\uparrow$0.9}\\
 & +AvisC \citep{woo2025don} & 79.1 {\color{red}$\downarrow$2.7} & 82.4 {\color{red}$\downarrow$1.1} & 71.8 {\color{red}$\downarrow$3.5} & 77.2 {\color{red}$\downarrow$1.5} & 64.4 {\color{red}$\downarrow$3.0} & 73.0 {\color{red}$\downarrow$0.7} & 71.8 {\color{red}$\downarrow$3.0} & 77.5 {\color{red}$\downarrow$1.1}\\
 & +VTI \citep{liu2025reducing} & 83.0 {\color{green!50!black}$\uparrow$1.2} & 83.6 {\color{green!50!black}$\uparrow$0.1} & 76.1 {\color{green!50!black}$\uparrow$0.8} & 79.5 {\color{green!50!black}$\uparrow$0.8} & 68.2 {\color{green!50!black}$\uparrow$0.8} & 74.6 {\color{green!50!black}$\uparrow$0.9} & 75.8 {\color{green!50!black}$\uparrow$1.0} & 79.2 {\color{green!50!black}$\uparrow$0.6}\\
 \rowcolor{blue!4}
 & +\textbf{RVSD (Ours)} & \textbf{87.5} {\color{green!50!black}$\uparrow$5.7} & \textbf{86.4} {\color{green!50!black}$\uparrow$2.9} & \textbf{85.4} {\color{green!50!black}$\uparrow$10.} & \textbf{84.5} {\color{green!50!black}$\uparrow$5.8} & \textbf{79.0} {\color{green!50!black}$\uparrow$11.} & \textbf{79.1} {\color{green!50!black}$\uparrow$5.4} & \textbf{84.0} {\color{green!50!black}$\uparrow$9.2} & \textbf{83.3} {\color{green!50!black}$\uparrow$4.7}\\
\midrule
\multirow{6}{*}{\textbf{LLaVA-NEXT}} & Vanilla & 83.8 {\color{gray}$\uparrow$0.0} & 83.0 {\color{gray}$\uparrow$0.0} & 81.4 {\color{gray}$\uparrow$0.0} & 80.8 {\color{gray}$\uparrow$0.0} & 73.2 {\color{gray}$\uparrow$0.0} & 74.5 {\color{gray}$\uparrow$0.0} & 79.5 {\color{gray}$\uparrow$0.0} & 79.4 {\color{gray}$\uparrow$0.0}\\
 & +VCD \citep{leng2024mitigating} & 84.8 {\color{green!50!black}$\uparrow$1.0} & 83.9 {\color{green!50!black}$\uparrow$0.9} & 81.5 {\color{green!50!black}$\uparrow$0.1} & 81.2 {\color{green!50!black}$\uparrow$0.4} & 74.7 {\color{green!50!black}$\uparrow$1.5} & 76.0 {\color{green!50!black}$\uparrow$1.5} & 80.3 {\color{green!50!black}$\uparrow$0.8} & 80.4 {\color{green!50!black}$\uparrow$1.0}\\
 & +M3ID \citep{favero2024multi} & 85.3 {\color{green!50!black}$\uparrow$1.5} & 84.3 {\color{green!50!black}$\uparrow$1.3} & 82.2 {\color{green!50!black}$\uparrow$0.8} & 81.6 {\color{green!50!black}$\uparrow$0.8} & 74.6 {\color{green!50!black}$\uparrow$1.4} & 75.7 {\color{green!50!black}$\uparrow$1.2} & 80.7 {\color{green!50!black}$\uparrow$1.2} & 80.5 {\color{green!50!black}$\uparrow$1.1}\\
 & +AvisC \citep{woo2025don} & 85.8 {\color{green!50!black}$\uparrow$2.0} & 84.4 {\color{green!50!black}$\uparrow$1.4} & 83.9 {\color{green!50!black}$\uparrow$2.5} & 82.8 {\color{green!50!black}$\uparrow$2.0} & 75.6 {\color{green!50!black}$\uparrow$2.4} & 76.6 {\color{green!50!black}$\uparrow$2.1} & 81.8 {\color{green!50!black}$\uparrow$2.3} & 81.3 {\color{green!50!black}$\uparrow$1.9}\\
 & +VTI \citep{liu2025reducing} & 84.8 {\color{green!50!black}$\uparrow$1.0} & 84.2 {\color{green!50!black}$\uparrow$1.2} & 80.4 {\color{red}$\downarrow$1.0} & 79.7 {\color{red}$\downarrow$1.1} & 72.8 {\color{red}$\downarrow$0.4} & 74.3 {\color{red}$\downarrow$0.2} & 79.3 {\color{red}$\downarrow$0.2} & 79.4 {\color{gray}$\uparrow$0.0}\\
 \rowcolor{blue!4}
 & +\textbf{RVSD (Ours)} & \textbf{90.1} {\color{green!50!black}$\uparrow$6.3} & \textbf{89.6} {\color{green!50!black}$\uparrow$6.6} & \textbf{88.6} {\color{green!50!black}$\uparrow$7.2} & \textbf{88.2} {\color{green!50!black}$\uparrow$7.4} & \textbf{81.8} {\color{green!50!black}$\uparrow$8.6} & \textbf{82.4} {\color{green!50!black}$\uparrow$7.9} & \textbf{86.8} {\color{green!50!black}$\uparrow$7.3} & \textbf{86.7} {\color{green!50!black}$\uparrow$7.3}\\
\midrule
\multirow{6}{*}{\textbf{Qwen-VL}} & Vanilla & 86.8 {\color{gray}$\uparrow$0.0} & 85.8 {\color{gray}$\uparrow$0.0} & 85.6 {\color{gray}$\uparrow$0.0} & 84.7 {\color{gray}$\uparrow$0.0} & 80.4 {\color{gray}$\uparrow$0.0} & 80.5 {\color{gray}$\uparrow$0.0} & 84.3 {\color{gray}$\uparrow$0.0} & 83.7 {\color{gray}$\uparrow$0.0}\\
 & +VCD \citep{leng2024mitigating} & 87.4 {\color{green!50!black}$\uparrow$0.6} & \textbf{86.6} {\color{green!50!black}$\uparrow$0.8} & 86.3 {\color{green!50!black}$\uparrow$0.7} & \textbf{85.1} {\color{green!50!black}$\uparrow$0.4} & 80.7 {\color{green!50!black}$\uparrow$0.3} & 80.8 {\color{green!50!black}$\uparrow$0.3} & 84.8 {\color{green!50!black}$\uparrow$0.5} & \textbf{84.2} {\color{green!50!black}$\uparrow$0.5}\\
 & +M3ID \citep{favero2024multi} & 87.1 {\color{green!50!black}$\uparrow$0.3} & 85.9 {\color{green!50!black}$\uparrow$0.1} & 85.9 {\color{green!50!black}$\uparrow$0.3} & 84.6 {\color{red}$\downarrow$0.1} & 80.5 {\color{green!50!black}$\uparrow$0.1} & 80.4 {\color{red}$\downarrow$0.1} & 84.5 {\color{green!50!black}$\uparrow$0.2} & 83.6 {\color{red}$\downarrow$0.1}\\
 & +AvisC \citep{woo2025don} & 84.7 {\color{red}$\downarrow$2.1} & 83.0 {\color{red}$\downarrow$2.8} & 83.9 {\color{red}$\downarrow$1.7} & 82.4 {\color{red}$\downarrow$2.3} & 78.1 {\color{red}$\downarrow$2.3} & 77.3 {\color{red}$\downarrow$3.2} & 82.2 {\color{red}$\downarrow$2.1} & 80.9 {\color{red}$\downarrow$2.8}\\
 & +VTI \citep{liu2025reducing} & 85.0 {\color{red}$\downarrow$1.8} & 85.3 {\color{red}$\downarrow$0.5} & 82.5 {\color{red}$\downarrow$3.1} & 83.3 {\color{red}$\downarrow$1.4} & 74.5 {\color{red}$\downarrow$5.9} & 75.0 {\color{red}$\downarrow$5.5} & 80.7 {\color{red}$\downarrow$3.6} & 81.2 {\color{red}$\downarrow$2.5}\\
 \rowcolor{blue!4}
 & +\textbf{RVSD (Ours)} & \textbf{87.4} {\color{green!50!black}$\uparrow$0.6} & 86.2 {\color{green!50!black}$\uparrow$0.4} & \textbf{87.7} {\color{green!50!black}$\uparrow$2.1} & 84.5 {\color{red}$\downarrow$0.2} & \textbf{82.8} {\color{green!50!black}$\uparrow$2.4} & \textbf{81.1} {\color{green!50!black}$\uparrow$0.6} & \textbf{86.0} {\color{green!50!black}$\uparrow$1.7} & 83.9 {\color{green!50!black}$\uparrow$0.2}\\
\bottomrule[1pt]
\end{tabular}
}
\captionsetup{font=small}
\caption{\label{POPE_Hall}Performance comparison on the POPE discrimination benchmark over the A-OKVQA dataset. The average column reports the mean accuracy and F1 scores across the random, popular, and adversarial splits. Green ({\color{green!50!black}$\bullet$}) and red ({\color{red}$\bullet$}) indicate improvements and degradations relative to the base model (sampling baseline), respectively, while ($\uparrow$/$\downarrow$) denotes the absolute change. Best results within each model block are highlighted in \textbf{bold}.}
\end{table*}

\begin{table*}[t]
\centering
\scriptsize
\setlength{\tabcolsep}{3.5pt}
\renewcommand{\arraystretch}{1.3}
\resizebox{\textwidth}{!}{
\begin{tabular}{llcccccccc}
\toprule[1pt]
\multirow{2}{*}{\textbf{Models}} & \multirow{2}{*}{\textbf{Methods}} & \multicolumn{2}{c}{\textbf{Random}} & \multicolumn{2}{c}{\textbf{Popular}} & \multicolumn{2}{c}{\textbf{Adversarial}} & \multicolumn{2}{c}{\textbf{Average}} \\
\cmidrule(lr){3-4} \cmidrule(lr){5-6} \cmidrule(lr){7-8} \cmidrule(lr){9-10}
& & \textbf{Acc (\%)$\uparrow$} & \textbf{F1 (\%)$\uparrow$} & \textbf{Acc (\%)$\uparrow$} & \textbf{F1 (\%)$\uparrow$} & \textbf{Acc (\%)$\uparrow$} & \textbf{F1 (\%)$\uparrow$} & \textbf{Acc (\%)$\uparrow$} & \textbf{F1 (\%)$\uparrow$}\\
\midrule
\multirow{6}{*}{\textbf{LLaVA-1.5}} & Vanilla & 81.6 {\color{gray}$\uparrow$0.0} & 83.5 {\color{gray}$\uparrow$0.0} & 73.1 {\color{gray}$\uparrow$0.0} & 77.5 {\color{gray}$\uparrow$0.0} & 68.0 {\color{gray}$\uparrow$0.0} & 74.5 {\color{gray}$\uparrow$0.0} & 74.2 {\color{gray}$\uparrow$0.0} & 78.5 {\color{gray}$\uparrow$0.0}\\
 & +VCD \citep{leng2024mitigating} & 82.2 {\color{green!50!black}$\uparrow$0.6} & 84.1 {\color{green!50!black}$\uparrow$0.6} & 71.5 {\color{red}$\downarrow$1.6} & 76.8 {\color{red}$\downarrow$0.7} & 67.6 {\color{red}$\downarrow$0.4} & 74.5 {\color{gray}$\uparrow$0.0} & 73.8 {\color{red}$\downarrow$0.4} & 78.5 {\color{gray}$\uparrow$0.0}\\
 & +M3ID \citep{favero2024multi} & 83.3 {\color{green!50!black}$\uparrow$1.7} & \textbf{84.5} {\color{green!50!black}$\uparrow$1.0} & 72.3 {\color{red}$\downarrow$0.8} & 77.1 {\color{red}$\downarrow$0.4} & 67.2 {\color{red}$\downarrow$0.8} & 74.0 {\color{red}$\downarrow$0.5} & 74.3 {\color{green!50!black}$\uparrow$0.1} & 78.5 {\color{gray}$\uparrow$0.0}\\
 & +AvisC \citep{woo2025don} & 79.0 {\color{red}$\downarrow$2.6} & 82.2 {\color{red}$\downarrow$1.3} & 67.4 {\color{red}$\downarrow$5.7} & 74.8 {\color{red}$\downarrow$2.7} & 64.1 {\color{red}$\downarrow$3.9} & 72.9 {\color{red}$\downarrow$1.6} & 70.2 {\color{red}$\downarrow$4.0} & 76.6 {\color{red}$\downarrow$1.9}\\
 & +VTI \citep{liu2025reducing} & 79.8 {\color{red}$\downarrow$1.8} & 82.3 {\color{red}$\downarrow$1.2} & 73.5 {\color{green!50!black}$\uparrow$0.4} & 78.0 {\color{green!50!black}$\uparrow$0.5} & 68.0 {\color{gray}$\uparrow$0.0} & 74.6 {\color{green!50!black}$\uparrow$0.1} & 73.8 {\color{red}$\downarrow$0.4} & 78.3 {\color{red}$\downarrow$0.2}\\
 \rowcolor{blue!4}
 & +\textbf{RVSD (Ours)} & \textbf{85.3} {\color{green!50!black}$\uparrow$3.7} & 83.8 {\color{green!50!black}$\uparrow$0.3} & \textbf{81.3} {\color{green!50!black}$\uparrow$8.2} & \textbf{80.3} {\color{green!50!black}$\uparrow$2.8} & \textbf{78.9} {\color{green!50!black}$\uparrow$11.} & \textbf{78.3} {\color{green!50!black}$\uparrow$3.8} & \textbf{81.8} {\color{green!50!black}$\uparrow$7.6} & \textbf{80.8} {\color{green!50!black}$\uparrow$2.3}\\
\midrule
\multirow{6}{*}{\textbf{LLaVA-NEXT}} & Vanilla & 83.1 {\color{gray}$\uparrow$0.0} & 82.5 {\color{gray}$\uparrow$0.0} & 78.5 {\color{gray}$\uparrow$0.0} & 78.5 {\color{gray}$\uparrow$0.0} & 73.3 {\color{gray}$\uparrow$0.0} & 74.5 {\color{gray}$\uparrow$0.0} & 78.3 {\color{gray}$\uparrow$0.0} & 78.5 {\color{gray}$\uparrow$0.0}\\
 & +VCD \citep{leng2024mitigating} & 83.4 {\color{green!50!black}$\uparrow$0.3} & 83.4 {\color{green!50!black}$\uparrow$0.9} & 78.2 {\color{red}$\downarrow$0.3} & 78.6 {\color{green!50!black}$\uparrow$0.1} & 74.2 {\color{green!50!black}$\uparrow$0.9} & 75.6 {\color{green!50!black}$\uparrow$1.1} & 78.6 {\color{green!50!black}$\uparrow$0.3} & 79.2 {\color{green!50!black}$\uparrow$0.7}\\
 & +M3ID \citep{favero2024multi} & 84.3 {\color{green!50!black}$\uparrow$1.2} & 84.3 {\color{green!50!black}$\uparrow$1.8} & 78.8 {\color{green!50!black}$\uparrow$0.3} & 78.9 {\color{green!50!black}$\uparrow$0.4} & 74.2 {\color{green!50!black}$\uparrow$0.9} & 75.4 {\color{green!50!black}$\uparrow$0.9} & 79.1 {\color{green!50!black}$\uparrow$0.8} & 79.5 {\color{green!50!black}$\uparrow$1.0}\\
 & +AvisC \citep{woo2025don} & 84.9 {\color{green!50!black}$\uparrow$1.8} & 83.5 {\color{green!50!black}$\uparrow$1.0} & 80.5 {\color{green!50!black}$\uparrow$2.0} & 79.5 {\color{green!50!black}$\uparrow$1.0} & 73.2 {\color{red}$\downarrow$0.1} & 74.3 {\color{red}$\downarrow$0.2} & 79.5 {\color{green!50!black}$\uparrow$1.2} & 79.1 {\color{green!50!black}$\uparrow$0.6}\\
 & +VTI \citep{liu2025reducing} & 83.3 {\color{green!50!black}$\uparrow$0.2} & 82.6 {\color{green!50!black}$\uparrow$0.1} & 78.6 {\color{green!50!black}$\uparrow$0.1} & 78.2 {\color{red}$\downarrow$0.3} & 73.2 {\color{red}$\downarrow$0.1} & 74.3 {\color{red}$\downarrow$0.2} & 78.4 {\color{green!50!black}$\uparrow$0.1} & 78.4 {\color{red}$\downarrow$0.1}\\
 \rowcolor{blue!4}
 & +\textbf{RVSD (Ours)} & \textbf{88.2} {\color{green!50!black}$\uparrow$5.1} & \textbf{87.2} {\color{green!50!black}$\uparrow$4.7} & \textbf{84.9} {\color{green!50!black}$\uparrow$6.4} & \textbf{84.3} {\color{green!50!black}$\uparrow$5.8} & \textbf{81.6} {\color{green!50!black}$\uparrow$8.3} & \textbf{81.4} {\color{green!50!black}$\uparrow$6.9} & \textbf{84.9} {\color{green!50!black}$\uparrow$6.6} & \textbf{84.3} {\color{green!50!black}$\uparrow$5.8}\\
\midrule
\multirow{6}{*}{\textbf{Qwen-VL}} & Vanilla & 81.3 {\color{gray}$\uparrow$0.0} & 79.2 {\color{gray}$\uparrow$0.0} & 75.9 {\color{gray}$\uparrow$0.0} & 74.9 {\color{gray}$\uparrow$0.0} & 75.5 {\color{gray}$\uparrow$0.0} & 74.4 {\color{gray}$\uparrow$0.0} & 77.6 {\color{gray}$\uparrow$0.0} & 76.2 {\color{gray}$\uparrow$0.0}\\
 & +VCD \citep{leng2024mitigating} & 82.0 {\color{green!50!black}$\uparrow$0.7} & 80.5 {\color{green!50!black}$\uparrow$1.3} & 75.9 {\color{gray}$\uparrow$0.0} & 75.6 {\color{green!50!black}$\uparrow$0.7} & 76.7 {\color{green!50!black}$\uparrow$1.2} & 76.2 {\color{green!50!black}$\uparrow$1.8} & 78.2 {\color{green!50!black}$\uparrow$0.6} & 77.4 {\color{green!50!black}$\uparrow$1.2}\\
 & +M3ID \citep{favero2024multi} & 82.4 {\color{green!50!black}$\uparrow$1.1} & 79.7 {\color{green!50!black}$\uparrow$0.5} & 76.8 {\color{green!50!black}$\uparrow$0.9} & 77.0 {\color{green!50!black}$\uparrow$2.1} & 77.1 {\color{green!50!black}$\uparrow$1.6} & 76.7 {\color{green!50!black}$\uparrow$2.3} & 78.8 {\color{green!50!black}$\uparrow$1.2} & 77.8 {\color{green!50!black}$\uparrow$1.6}\\
 & +AvisC \citep{woo2025don} & 80.5 {\color{red}$\downarrow$0.8} & 77.8 {\color{red}$\downarrow$1.4} & 74.2 {\color{red}$\downarrow$1.7} & 72.3 {\color{red}$\downarrow$2.6} & 75.5 {\color{gray}$\uparrow$0.0} & 73.4 {\color{red}$\downarrow$1.0} & 76.7 {\color{red}$\downarrow$0.9} & 74.5 {\color{red}$\downarrow$1.7}\\
 & +VTI \citep{liu2025reducing} & 82.0 {\color{green!50!black}$\uparrow$0.7} & \textbf{82.4} {\color{green!50!black}$\uparrow$3.2} & 76.5 {\color{green!50!black}$\uparrow$0.6} & 77.9 {\color{green!50!black}$\uparrow$3.0} & 71.0 {\color{red}$\downarrow$4.5} & 74.5 {\color{green!50!black}$\uparrow$0.1} & 76.5 {\color{red}$\downarrow$1.1} & 78.3 {\color{green!50!black}$\uparrow$2.1}\\
 \rowcolor{blue!4}
 & +\textbf{RVSD (Ours)} & \textbf{84.3} {\color{green!50!black}$\uparrow$3.0} & 81.7 {\color{green!50!black}$\uparrow$2.5} & \textbf{82.7} {\color{green!50!black}$\uparrow$6.8} & \textbf{80.3} {\color{green!50!black}$\uparrow$5.4} & \textbf{79.9} {\color{green!50!black}$\uparrow$4.4} & \textbf{77.8} {\color{green!50!black}$\uparrow$3.4} & \textbf{82.3} {\color{green!50!black}$\uparrow$4.7} & \textbf{79.9} {\color{green!50!black}$\uparrow$3.7}\\
\bottomrule[1pt]
\end{tabular}
}
\captionsetup{font=small}
\caption{\label{POPE_Hall_GQA}Performance comparison on the POPE discrimination benchmark over the GQA dataset. The average column reports the mean accuracy and F1 scores across the random, popular, and adversarial splits. Green ({\color{green!50!black}$\bullet$}) and red ({\color{red}$\bullet$}) indicate improvements and degradations relative to the base model (sampling baseline), respectively, while ($\uparrow$/$\downarrow$) denotes the absolute change. Best results within each model block are highlighted in \textbf{bold}.}
\end{table*}

\section{More Experiments and Results}
\label{sec:appendix}

\noindent\textbf{Results on POPE A-OKVQA and GQA.}
To further evaluate the cross-dataset generalization of RVSD, we conduct additional experiments on the POPE A-OKVQA and GQA subsets, as shown in Tables~\ref{POPE_Hall} and~\ref{POPE_Hall_GQA}. Compared with POPE-MSCOCO (refer to Table~\ref{tab:chair_pope_compare}), these datasets contain denser scenes and stronger object co-occurrence biases, making the popular and adversarial splits more challenging. RVSD achieves the best accuracy across all splits on all three backbones and the best F1 scores on the two LLaVA backbones, with the largest improvements observed on the adversarial split. In particular, RVSD improves Accuracy/F1 by 11.6/5.4 on LLaVA-1.5 A-OKVQA and by 10.9/3.8 on LLaVA-1.5 GQA, demonstrating strong robustness against misleading linguistic priors. On Qwen-VL, RVSD still obtains the highest accuracy on every split, improving the adversarial accuracy by 2.4 on A-OKVQA and by 4.4 on GQA, while its F1 scores remain comparable to the strongest contrastive baseline. In contrast, existing baselines provide only limited or inconsistent improvements, while AvisC notably degrades the average performance on both A-OKVQA and GQA. These results demonstrate that the effectiveness of RVSD generalizes consistently across different POPE datasets and is not dependent on a specific data distribution.

\section{Detailed Experimental Setup}
\label{app:exp_setup}

\subsection{Datasets and Benchmarks}
\label{app:dataset}

We provide additional details for the five benchmarks used in the main paper, all of which are publicly released and standard in the VH literature.

\smallskip\noindent\textbf{POPE}~\citep{li2023evaluating} evaluates object-level hallucination through binary ``Yes/No'' existence questions and reports accuracy and F1 scores under three sampling strategies, i.e., random, popular, and adversarial splits. The popular split samples frequently co-occurring object categories. In contrast, the adversarial split selects objects with strong co-occurrence priors relative to the image, making it particularly challenging for VH mitigation. We evaluate POPE on three data sources, namely MSCOCO~\citep{rohrbach2018object}, A-OKVQA, and GQA, each containing 3,000 yes/no queries.

\smallskip\noindent\textbf{MME Hallucination Subset}~\citep{fu2026mme} evaluates hallucination through four probing tasks, including object-level probes (Existence, Count) and attribute-level probes (Position, Color). Each task contains 30 binary questions and is evaluated using Acc/Acc+, where Acc+ requires both yes/no questions associated with the same image to be answered correctly. The total score is computed as the sum of the four subtask scores.

\smallskip\noindent\textbf{AMBER}~\citep{wang2023amber} contains both a discrimination subset and a generative subset. The discrimination subset uses binary yes/no questions and is evaluated with accuracy and F1 scores, while the generative subset evaluates open-ended captioning using CHAIR, HalRate, and a holistic Score that jointly measures caption coverage and hallucination penalty. AMBER covers existence-, attribute-, and relation-level hallucinations, with a particular emphasis on long-form caption generation.

\smallskip\noindent\textbf{CHAIR}~\citep{rohrbach2018object} evaluates object hallucination in generated captions by extracting mentioned objects and matching them against MSCOCO ground-truth annotations. It reports two metrics, which are given by
\begin{equation}
\small
\mathrm{CHAIR}_I=\frac{|\{\mathrm{hallucinated objects}\}|}{|\{\mathrm{all mentioned objects}\}|},
\end{equation}
\begin{equation}
\small
\mathrm{CHAIR}_S=\frac{|\{\mathrm{captions w/ hallucination}\}|}{|\{\mathrm{all captions}\}|}.
\end{equation}
Following~\citep{liu2025reducing}, we randomly sample 500 MSCOCO Val2014 images and prompt the model with ``\textit{Please describe this image in detail}'' to elicit long-form responses.

\smallskip\noindent\textbf{MM-Vet}~\citep{pmlr-v235-yu24o} consists of 218 evaluation samples designed to probe six vision-language capabilities: Recognition (Rec), OCR, Knowledge (Know), Generation (Gen), Spatial reasoning (Spat), and Math. We follow the official evaluation protocol and use GPT-4 as the judge to assign a continuous score for each response, with the total score computed as the mean across all dimensions.

\subsection{Baselines}
\label{app:baselines}

We compare RVSD against four representative training-free decoders spanning three paradigms. All baselines follow the recommended hyperparameters and are evaluated under the same decoding budget as RVSD for a fair comparison.

\smallskip\noindent\textbf{VCD} (Visual Contrastive Decoding)~\citep{leng2024mitigating} introduces a perturbed visual input by adding diffusion-style Gaussian noise and performs contrastive decoding between the clean and corrupted visual conditions, thereby suppressing unimodal language priors.

\smallskip\noindent\textbf{M3ID} (Multi-Modal Mutual-Information Decoding)~\citep{favero2024multi} reweights output logits using a mutual-information-based score between visual and textual modalities, aiming to enhance visual grounding during generation.

\smallskip\noindent\textbf{AvisC} (Attentional Vision Calibration)~\citep{woo2025don} identifies high-attention ``blind tokens'' that dominate cross-modal attention and recalibrates predictions via dual forward passes with and without these tokens.

\smallskip\noindent\textbf{VTI} (Visual Token Intervention)~\citep{liu2025reducing} steers latent representations in the large language model hidden space along a pre-computed direction associated with improved visual grounding.

\subsection{Models}
\label{app:models}

We instantiate RVSD on three LVLMs drawn from two distinct architectural families. \textbf{LLaVA-1.5-7B}~\citep{liu2024improved} adopts CLIP-ViT-L/14-336px as the visual encoder, a two-layer MLP as the cross-modal projector, and Vicuna-7B-v1.5 as the language backbone, producing 576 visual tokens per image. \textbf{LLaVA-NEXT-7B}~\citep{liu2024llavanext} extends LLaVA-1.5 with the AnyRes mechanism, which partitions the input image into up to four high-resolution sub-images, increasing the number of visual tokens to as many as 2,880 and substantially amplifying the burden on visual grounding. \textbf{Qwen-VL}~\citep{bai2023qwen} instead couples OpenCLIP-ViT-bigG with a single-layer position-aware cross-attention adapter that compresses the visual feature map into a fixed set of 256 visual tokens, and adopts Qwen-7B as the language backbone, differing from the LLaVA series in both the vision-language interface and the language backbone. Overall, the backbones cover standard and high-resolution settings as well as both MLP-projection and cross-attention resampler interfaces, and demonstrate that RVSD is model-agnostic.

\begin{table*}[t]
\centering
\small
\renewcommand{\arraystretch}{1.25}
\setlength{\tabcolsep}{6pt}
\resizebox{\textwidth}{!}{
\begin{tabular}{clccc}
\toprule[1.3pt]
\rowcolor{gray!8}
\textbf{Symbols} & \textbf{Descriptions} & \textbf{LLaVA-1.5} & \textbf{LLaVA-NEXT} & \textbf{Qwen-VL} \\
\midrule
$k$              & Sparse-set budget          & 192                                 & Scaled by \texttt{image\_shape} & 192 (tier) $\to$ [133, 89, 52] \\
$m$              & Major textual tokens     & $\lfloor|y_{<t}|/\text{2}\rfloor$         & $\lfloor|y_{<t}|/\text{2}\rfloor$ & $\lfloor|y_{<t}|/\text{2}\rfloor$ \\
$\tau$           & Saliency softmax temperature & 1.0                                & 1.0 & 1.0 \\
$\eta$           & Temporal smoothing coefficient & 0                                & 0 & 0 \\
$\mathcal{L}_s$  & Pruning layer set  & $\{\text{2, 6, 15}\}$                       & $\{\text{2, 6, 15}\}$ & $\{\text{2, 6, 15}\}$ \\
$\gamma$         & Entropy trigger threshold    & 0.75                               & 0.5 & 0.75 \\
$[l_{\mathrm{s}},l_{\mathrm{e}}]$ & SSVR layer scanning window & [6,27]              & [5,12] & [5,16] \\
$K$              & Retrieval top-$K$ bound      & 128                                & 7 & 57 \\
$\alpha$         & Injection ratio              & 0.2                                & 0.09 & 0.2 \\
\bottomrule[1.3pt]
\end{tabular}
}
\captionsetup{font=small}
\caption{\label{tab:hparam}Default hyperparameter settings of RVSD used across all experiments unless otherwise specified.}
\end{table*}

\section{Implementation Details}
\label{app:impl}

\noindent\textbf{Hyperparameter Settings of RVSD.}
RVSD is implemented in PyTorch on top of the official LLaVA codebase and introduces no additional trainable parameters. Table~\ref{tab:hparam} summarizes the default hyperparameter configuration used across all experiments unless otherwise specified. We adopt a unified pruning layer set $\mathcal{L}_s={\text{2, 6, 15}}$ for all three backbones, where the visual-token bank is progressively pruned at these layers. For LLaVA-1.5-7B, the sparse-set budget is set to $k=\text{192}$, approximately one third of $N_v=\text{576}$, resulting in a 67\% reduction in visual tokens. For LLaVA-NEXT-7B, the budget is rescaled according to the image shape to maintain a consistent retention ratio under the AnyRes high-resolution setting. For Qwen-VL, whose cross-attention adapter emits only $N_v=\text{256}$ visual tokens, the budget starts from $k=\text{192}$ and is progressively tightened to [133, 89, 52] at the three pruning layers, yielding a final retention ratio comparable to that of LLaVA-1.5. The number of major textual tokens is defined as $m=\lfloor|y{<t}|/\text{2}\rfloor$, which selects the most recent half of the generated prefix for visual-saliency aggregation and ensures a length-adaptive allocation consistent with the Top-K formulation in the main paper. Temporal smoothing in Eq.~(\ref{eq:smooth}) is disabled by setting $\eta=\text{0}$, i.e., $\tilde{s}^t=s^t$. For SSVR, the entropy threshold $\gamma$, the intermediate-layer window $[l_{\mathrm{s}},l_{\mathrm{e}}]$, the retrieval bound $K$, and the injection ratio $\alpha$ are tuned per backbone to accommodate differences in visual-token granularity across LLaVA-1.5 (single 24$\times$24 grid), LLaVA-NEXT (multi-granularity AnyRes grids), and Qwen-VL (256 resampled tokens without an explicit spatial grid).

\smallskip\noindent\textbf{Decoding Protocol.}
Following~\citep{leng2024mitigating,liu2025reducing}, all methods, including RVSD and baselines, are evaluated under a unified decoding setup using nucleus sampling with temperature $T=\text{1.0}$, top-$p=\text{0.9}$, no repetition penalty, and batch size 1 for a fair comparison. The generation budget, i.e., the maximum number of new tokens, is benchmark-dependent: It is set to 2 for binary yes/no tasks in POPE and the MME hallucination subset, and to 1,024 for open-ended generation benchmarks including AMBER (generative), MM-Vet, and CHAIR. For the inference latency evaluation in Table~\ref{tab:latency}, we follow the protocol specified in the table caption, setting the maximum number of new tokens to 32 with batch size 1.

\smallskip\noindent\textbf{Implementation Details.}
We conduct all experiments on a single NVIDIA RTX A6000 GPU (48 GB) with NVIDIA driver 535.183.01 and CUDA 12.2 on Ubuntu 22.04. The software stack includes PyTorch 2.0 and Hugging Face Transformers 4.40.

\section{Hyperparameter Sensitivity Analysis}
\label{app:sensitivity}

\begin{figure}[t]
\centering
\includegraphics[width=0.43\textwidth]{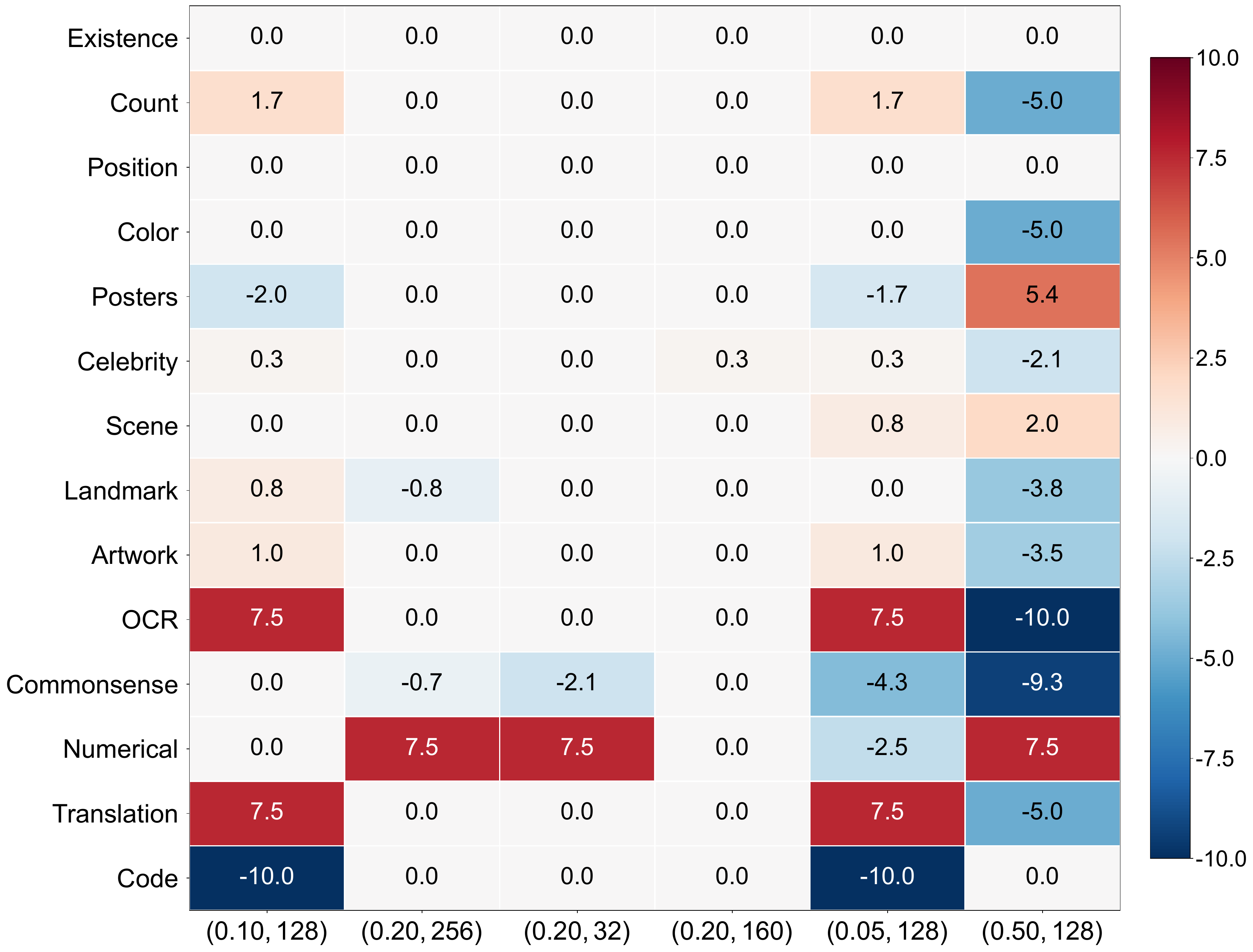}
\captionsetup{font=small}
\caption{\label{fig:mme_ablation}Subtask-level $\Delta$ heatmap on MME under non-baseline $(\alpha, k_c)$ configurations, where $\Delta = \mathrm{score(cfg)} - \mathrm{score(base)}$ is computed relative to the baseline setting $(\alpha=\text{0.20}, k_c=\text{128})$.}
\end{figure}

\noindent\textbf{Subtask-Level Sensitivity on MME.}
We first conduct a one-factor-at-a-time analysis around the baseline $(\alpha=\text{0.20}, k_c=\text{128})$ on MME with LLaVA-1.5, where $\alpha$ denotes the retrieval injection ratio and $k_c$ the top-$k$ retrieval size. As shown in Fig.~\ref{fig:mme_ablation}, $\alpha$ is the primary driver of subtask-level variations: Small values of $\alpha$ (0.05/0.10) consistently improve OCR and translation by up to +7.5, whereas a larger value ($\alpha=\text{0.50}$) leads to broad degradations across OCR, commonsense, counting, and color reasoning. In contrast, varying $k_c$ within [32, 256] yields negligible changes across all subtasks. This analysis identifies $(\alpha=\text{0.10}, k_c=\text{128})$ as a stable operating point and further reveals that the aggregate MME score can obscure task-specific regressions, particularly in reasoning-intensive categories.

\noindent\textbf{Hyperparameter Sensitivity.}
To complement the subtask-level analysis above, we further investigate the sensitivity of RVSD to its four key hyperparameters on the POPE-MSCOCO adversarial split with LLaVA-1.5, which is the most discriminative setting for hallucination evaluation. We perform a one-factor-at-a-time analysis, varying each hyperparameter while keeping the others fixed at their default values in Table~\ref{tab:hparam}, and the results are reported in Table~\ref{tab:hparam_sens}.

\begin{table}[t]
\centering
\renewcommand{\arraystretch}{1.2}
\resizebox{\columnwidth}{!}{
\begin{tabular}{lccc}
\toprule[1.5pt]
\rowcolor{gray!8}
\textbf{Hyperparameters} & \textbf{Setting} & \textbf{Acc (\%) $\uparrow$} & \textbf{F1 (\%) $\uparrow$} \\
\midrule
\multirow{4}{*}{Injection ratio $\alpha$}
 & 0.05 & 84.7          & 83.4 \\
 & 0.10 & 84.7          & 83.4 \\
 & 0.20 & \textbf{84.7} & \textbf{83.5} \\
 & 0.50 & 84.4          & 83.2 \\
\midrule
\multirow{4}{*}{Retrieval bound $K$}
 & 32   & 84.6          & 83.4 \\
 & 64   & 84.6          & 83.4 \\
 & 128  & \textbf{84.7} & \textbf{83.5} \\
 & 256  & 84.6          & 83.4 \\
\midrule
\multirow{3}{*}{Entropy threshold $\gamma$}
 & 0.3  & \textbf{84.7} & 83.4 \\
 & 0.5  & \textbf{84.7} & 83.4 \\
 & 0.7  & 84.6          & 83.4 \\
\midrule
\multirow{4}{*}{Sparse budget $k$}
 & \phantom{0}96   & 83.0          & 81.1 \\
 & 192             & 84.7          & 83.5 \\
 & 288             & \textbf{85.0} & \textbf{84.1} \\
 & 576 ($=\!N_v$)  & \textbf{85.0} & \textbf{84.1} \\
\bottomrule[1.5pt]
\end{tabular}
}
\captionsetup{font=small}
\caption{\label{tab:hparam_sens}Hyperparameter sensitivity analysis on the POPE-MSCOCO Adversarial split with LLaVA-1.5. The best result in each block is highlighted in \textbf{bold}.}
\end{table}

\smallskip\noindent\textbf{Effect of Injection Ratio $\alpha$.}
RVSD is robust to the injection ratio $\alpha$ in the practical range [0.05, 0.20], where Accuracy/F1 remains stable at 84.7/83.4-84.7/83.5. Increasing $\alpha$ to 0.50 over-injects retrieved evidence relative to the residual-stream magnitude, leading to a mild degradation to 84.4/83.2. This behavior is consistent with the amplitude-alignment design in Eq.~(\ref{eq:adapter_w}), where the retrieval branch is intended to re-anchor rather than overwrite the host representation.

\smallskip\noindent\textbf{Effect of Retrieval Bound $K$.}
Accuracy/F1 scores vary by less than 0.1 across the retrieval bound $K \in \{\text{32, 64, 128, 256}\}$, indicating strong robustness of the softmax-weighted aggregation in Eq.~(\ref{eq:retrieval}). Irrelevant entries in the deferred memory bank receive near-zero weights and are effectively suppressed, such that increasing $K$ neither introduces noise nor yields additional benefit.

\smallskip\noindent\textbf{Effect of Entropy Threshold $\gamma$.}
RVSD is stable across $\gamma \in {\text{0.3, 0.5, 0.7}}$, with the accuracy score remaining at 84.7 for the two lower thresholds and decreasing by only 0.1 at $\gamma=\text{0.7}$. This suggests that, on the POPE adversarial split, the SSVR gate activates on a well-separated subset of uncertain steps, and moderate variations in $\gamma$ do not materially affect which steps are re-anchored.

\smallskip\noindent\textbf{Effect of Sparse Budget $k$.}
The sparse budget $k$ is the most influential among the four hyperparameters. A small budget ($k=\text{96}$, approximately one-sixth of $N_v$) removes too much fine-grained visual evidence and reduces Accuracy/F1 to 83.0/81.1. Increasing $k$ to 288 or 576 slightly improves performance to 85.0/84.1. The default setting ($k=\text{192}$) achieves competitive Accuracy/F1 (84.7/83.5) while maintaining a substantially lower attention cost than $k=\text{576}$, and is therefore adopted as the efficiency–faithfulness trade-off point reported in Table~\ref{tab:latency}. This trend empirically supports the claim in Section~\ref{sec:dyn_sparse} that visual grounding in RVSD is primarily driven by semantics-directed selection rather than the sheer number of retained tokens.

\section{Algorithmic Complexity}
\label{app:complexity}

We analyze the asymptotic overhead introduced by RVSD over vanilla decoding. Let $D$ denote the hidden dimension, $N_v$ the number of visual tokens, $t$ the current decoding step, and $k$ the sparse-set budget. The dominant cost of vanilla decoding at step $t$ arises from per-layer self-attention, with complexity $\mathcal{O}((t+N_v)^2 D)$. In RVSD, this is reduced to $\mathcal{O}((t+k)^2 D)$ through the sparse set $\mathcal{S}_t$, leading to computational savings when $k \ll N_v$.

The additional overhead of RVSD consists of three components. First, the textual saliency computation in Eq.~(\ref{eq:saliency}) and the relevance scoring in Eq.~(\ref{eq:relevance}) incur costs of $\mathcal{O}(t)$ and $\mathcal{O}(m N_v |\mathcal{L}_s|)$, respectively. Both are linear and negligible compared with the quadratic attention term. Second, SSVR retrieval in Eq.~(\ref{eq:retrieval}) requires $\mathcal{O}(|\mathcal{P}_t| D)$ for similarity computation and $\mathcal{O}(|\mathcal{P}_t| \log K)$ for top-$K$ selection, which remains dominated by self-attention. Third, the residual-stream adapter in Eq.~(\ref{eq:adapter_w}) and the fusion step in Eq.~(\ref{eq:fusion}) incur $\mathcal{O}(K D)$ per triggered step. However, the entropy-based gate in Eq.~(\ref{eq:uncertainty}) activates retrieval only on a small subset of steps. Overall, the additional per-step cost is bounded by $\mathcal{O}(N_v D)$ and is negligible relative to attention. Empirically, Table~\ref{tab:latency} further shows that RVSD reduces per-token latency, maximum memory, TFLOPs, and TTFT compared with vanilla decoding, a profile not achieved by previous decoding-side methods for VH mitigation.

\section{License of Artifacts}
\label{app:license}

We use all datasets and models strictly for academic research in accordance with their original licenses and terms. MSCOCO is released under the Creative Commons Attribution 4.0 License, and POPE is released under the MIT License. AMBER, MME, and MM-Vet are used under their respective official non-commercial academic research licenses. LLaVA-1.5 and LLaVA-NEXT are released under the Apache 2.0 License, while their LLaMA-2/Vicuna backbones are subject to the Meta LLaMA-2 community license. Qwen-VL is released under the Tongyi Qianwen License, which allows use for academic research. We do not redistribute any of these resources beyond the permissions granted by their licenses, and we will release our RVSD implementation under a permissive open-source license upon publication.

\section{Case Study}
  \label{app:case_study}

\begin{figure*}[p]
      \centering
      \includegraphics[width=\textwidth,height=0.95\textheight,keepaspectratio]{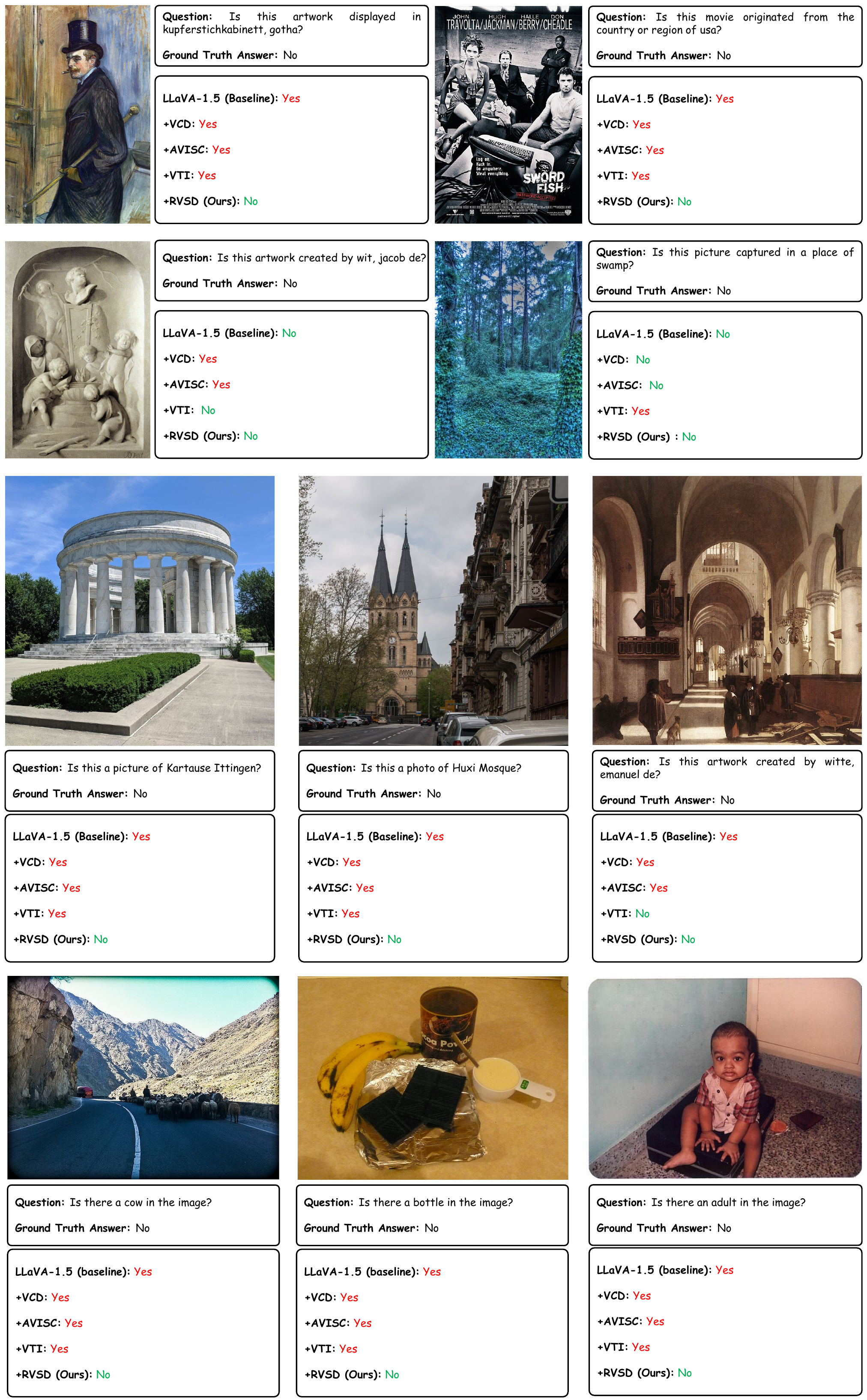}
      \captionsetup{font=small}
      \caption{\label{fig:case_study} Qualitative case studies of RVSD compared with the vanilla decoder and baseline methods. Hallucinated content is highlighted in {\color{red}red}, while faithful content grounded in the input image is highlighted in {\color{green!50!black}green}. RVSD consistently produces visually faithful descriptions across diverse scenes.}
  \end{figure*}

\begin{figure*}[p]
      \centering
      \includegraphics[width=\textwidth,height=0.95\textheight,keepaspectratio]{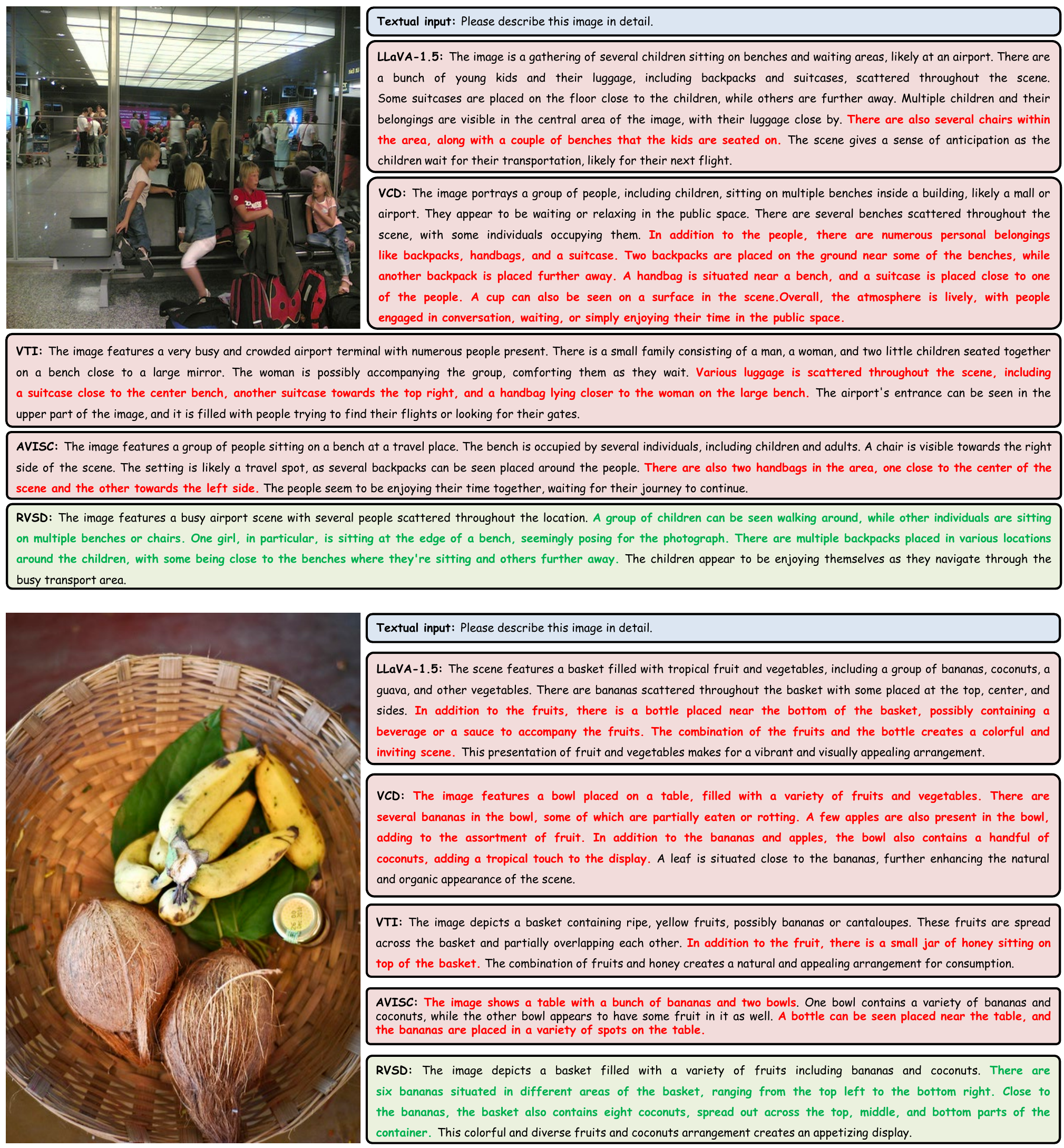}
      \captionsetup{font=small}
      \caption{\label{fig:case_study4} Additional qualitative case studies of RVSD on more challenging long-form generation samples. Hallucinated content is highlighted in {\color{red}red}, while faithful content grounded in the input image is highlighted in {\color{green!50!black}green}. RVSD continues to deliver visually faithful generations even under denser scenes and finer-grained semantic dependencies, where baseline decoders frequently exhibit object-, attribute-, and relation-level hallucinations.}
  \end{figure*}

  To provide an intuitive understanding of how RVSD mitigates VHs during generation, we present qualitative case studies in Figs.~\ref{fig:case_study} and~\ref{fig:case_study4}. Compared with the vanilla decoder and representative sparse decoding baselines, RVSD produces captions that are more faithfully grounded in the visual content, successfully suppressing hallucinated objects, attributes, and spatial relations that are absent from the input image. As shown in Fig.~\ref{fig:case_study}, RVSD consistently mitigates object- and attribute-level hallucinations across diverse scenes, while Fig.~\ref{fig:case_study4} further demonstrates its effectiveness under more challenging long-form generation settings with denser visual content and finer-grained semantic dependencies. These cases collectively demonstrate that the semantics-directed token selection effectively preserves visually critical evidence, while the SSVR mechanism re-anchors decoding through on-demand cross-modal retrieval whenever visual grounding becomes insufficient.

\section{Usage of AI Assistant}
\label{app:ai_usage}

We used GPT-4o exclusively for the language polishing of this manuscript, including grammar correction and sentence-level rephrasing. All generated outputs were manually reviewed and edited to ensure technical accuracy. No part of the research ideation, method design, experimental analysis, or interpretation of the results involved AI assistance.

\end{document}